\documentclass[journal]{IEEEtran}
\usepackage{amsmath,amsfonts}
\usepackage{array}
\usepackage{textcomp}
\usepackage{stfloats}
\usepackage{url}
\usepackage{verbatim}
\usepackage{graphicx}
\def\BibTeX{{\rm B\kern-.05em{\sc i\kern-.025em b}\kern-.08em
    T\kern-.1667em\lower.7ex\hbox{E}\kern-.125emX}}
\usepackage{balance}

\usepackage{algorithm}
\usepackage{algpseudocode}
\algrenewcommand\algorithmicrequire{\textbf{Input:}}
\algrenewcommand\algorithmicensure{\textbf{Output:}}

\usepackage{cite}
\usepackage{tikz}
\usepackage{pgfplots}
\usepackage{xcolor}
\pgfplotsset{compat=1.18}
\usetikzlibrary{positioning}
\usetikzlibrary{decorations.pathreplacing}
\usetikzlibrary{calc}
\usetikzlibrary{positioning,shapes,shadows,arrows}

\usepackage{pifont}
\newcommand{\cmark}{\ding{51}}%
\newcommand{\xmark}{\ding{55}}%
\usepackage{multirow}
\usepackage{bbm}
\usepackage{enumitem}
\usepackage{subcaption}
\usepackage{bm}
\usepackage[font=small]{caption}
\usepackage{threeparttable}

\pgfplotsset{
every axis/.append style={
  axis line style={->}, 
  legend style={font=\scriptsize},
  label style={font=\scriptsize},
  title style={font=\scriptsize},
  tick label style={font=\scriptsize},
  }
}

\makeatletter
\let\NAT@parse\undefined
\makeatother
\usepackage{hyperref}

\begin{document}
\title{Object-Centric Kinodynamic Planning for Nonprehensile Robot Rearrangement Manipulation}

\author{Kejia Ren$^1$, 
        Gaotian Wang$^1$, 
        Andrew S. Morgan$^2$, 
        Lydia E. Kavraki$^1$, 
        and Kaiyu Hang$^1$
\thanks{This work was supported by the U.S. National Science Foundation (NSF) under grants FRR-2133110 and FRR-2240040 (KR, GW, and KH), and grant CCF-2336612 (LEK).}
\thanks{$^1$Kejia Ren, Gaotian Wang, Lydia E. Kavraki, and Kaiyu Hang are with the Department of Computer Science, Rice University, Houston, TX 77005, USA (email: kr43@rice.edu; gw23@rice.edu; kavraki@rice.edu; kaiyu.hang@rice.edu).}
\thanks{$^{2}$Andrew S. Morgan is with the RAI Institute, Cambridge, MA 02142, USA (e-mail: andy@rai-inst.com).}
}

\maketitle

\begin{abstract}
Nonprehensile actions such as pushing are crucial for addressing multi-object rearrangement problems.
Many traditional methods generate \emph{robot-centric} actions,
which differ from intuitive human strategies and are typically inefficient.
To this end, 
we adopt an \emph{object-centric} planning paradigm and propose a unified framework for addressing a range of large-scale, physics-intensive nonprehensile rearrangement problems challenged by modeling inaccuracies and real-world uncertainties.
By assuming each object can actively move without being driven by robot interactions, our planner first computes desired object motions, which are then realized through robot actions generated online via a closed-loop pushing strategy.
Through extensive experiments and in comparison with state-of-the-art baselines in both simulation and on a physical robot, 
we show that our \emph{object-centric} planning framework can generate more intuitive and task-effective robot actions with significantly improved efficiency.
In addition, we propose a benchmarking protocol to standardize and facilitate future research in nonprehensile rearrangement.
\end{abstract}

\section{Introduction}
\label{sec:intro}

Rearrangement of multiple objects, which refers to reconfiguring objects into certain desired states, is generally required for various practical manipulation tasks such as singulation for object retrieval~\cite{dogar2011framework, lee2019efficient}, obstacle clearance for navigation~\cite{stilman2005navigation, nieuwenhuisen2008effective, van2010path}, multi-object sorting~\cite{gupta2012using}, etc.
It is important in many real-world applications, 
such as warehouse sorting~\cite{talia2023pushr}, recycling lines~\cite{han2021toward}, and home cleaning~\cite{szot2021habitat};
and often serves as a key step in downstream manipulation pipelines such as bin picking~\cite{nieuwenhuisen2013mobile} and packing~\cite{shome2019towards}.
As an essential manipulation skill of robots, rearrangement is enabled by planning the robot's motion with necessary constraints such as collision avoidance and the robot's kinematics,
which is proven NP-hard~\cite{wilfong1988motion}.
Traditional planning methods allow only a pick-and-place type of prehensile action to move one object at a time~\cite{simeon2004manipulation, stilman2007manipulation}.
Although certain optimality can be achieved, prehensile action-based rearrangement can be ineffective or even infeasible since the object geometry (e.g., size or shape) or other environment-relevant constraints (e.g., limited free space) can make the objects not graspable.
 
\begin{figure}[t]
    \centering
    \includegraphics[width=0.95\columnwidth]{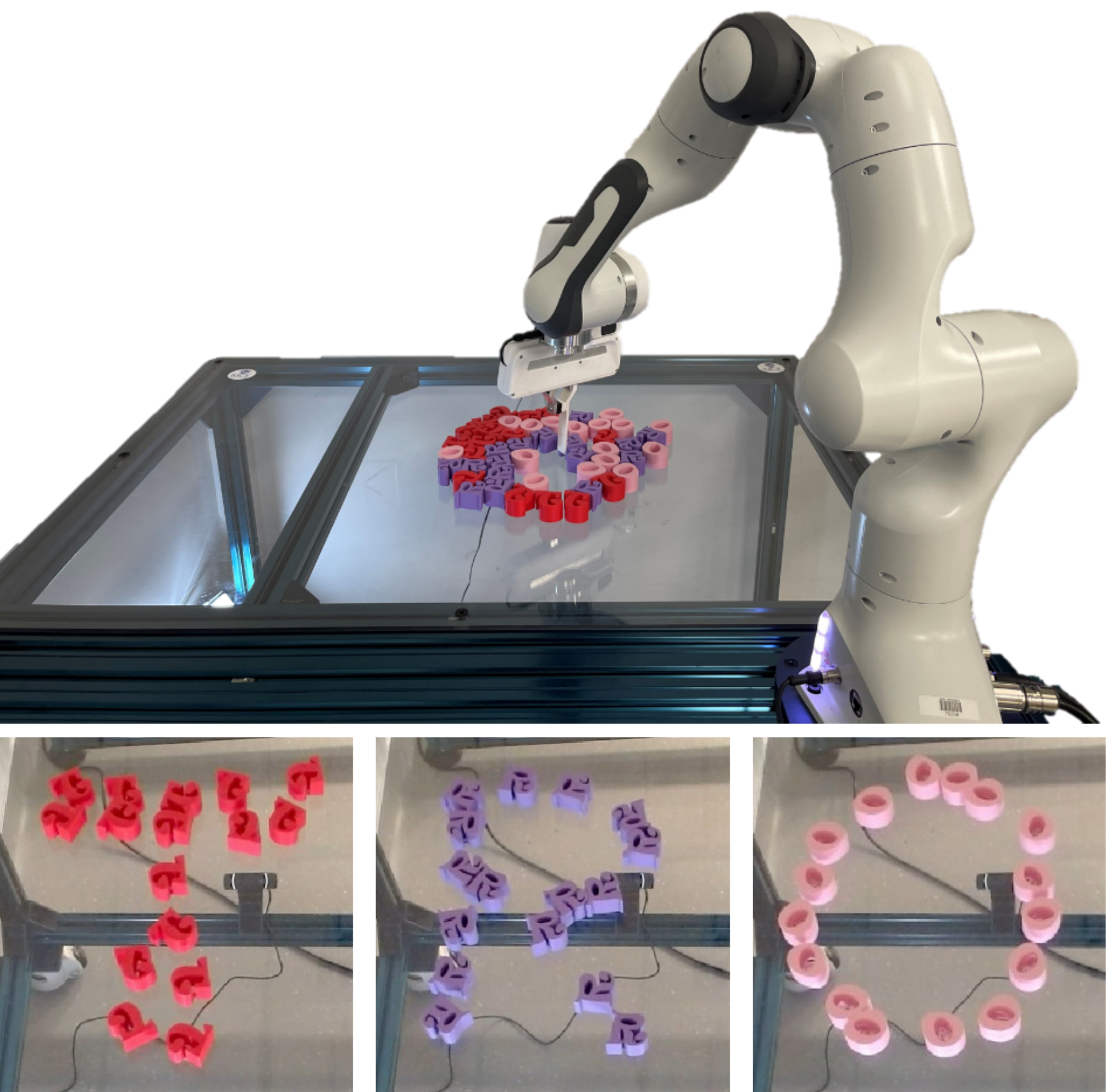}
    \caption{Through object-centric planning, our framework is able to efficiently rearrange multiple movable objects of different shapes to accomplish various tasks.
    In the scene, ``T'', ``R'', and ``O'' letter-shaped objects are rearranged to form the abbreviation ``TRO''.}
    \label{fig:first}
\end{figure}

Thereafter, nonprehensile actions such as pushing have been investigated.
Nonprehensile action-based methods can generate more diverse and effective solutions to object rearrangement, by modeling the interaction physics between the robot and objects and allowing concurrent manipulation of multiple objects~\cite{ruggiero2018nonprehensile}.
However, nonprehensile rearrangement planning is challenging due to the sophisticated physics modeling and the high-dimensional problem space it entails.
Furthermore, the inevitable modeling inaccuracy of the involved physics makes nonprehensile solutions less robust to real-world uncertainties and causes task failures.
To this end, developing an efficient and also reactive manipulation planner for nonprehensile multi-object rearrangement problems is highly desired.
In this work, we develop a manipulation planner for large-scale nonprehensile rearrangement problems where the object clutters are highly packed and concurrent interactions between objects are inevitably common, as exemplified in Fig.~\ref{fig:first}.

\emph{Robot-Centric vs. Object-Centric Planning.}
Traditional nonprehensile rearrangement planners~\cite{haustein2015kinodynamic, king2015nonprehensile} are mostly \emph{robot-centric}:
During planning, robot actions (e.g., joint-space control) are sampled considering only robot-relevant geometric and kinematic constraints.
The resulting object motions are predicted after sampling and used to select promising samples for execution.
Such \emph{robot-centric} strategies can be inefficient, as planners often spend unnecessary time sampling and evaluating actions that fail to make meaningful interactions with objects.
To this end, \emph{object-centric} actions, generated with object-relevant intent~\cite{king2016rearrangement}, have been incorporated.
Early work~\cite{sundaram2001disassembly, krontiris2014rearranging} treat the objects (or parts) to be manipulated as actively movable robots, but focused on geometric or prehensile manipulation planning problems,
without the ability to integrate physics-involved nonprehensile actions.
To better target action generation towards interactions with objects, prior works either define \emph{object-centric} action primitives~\cite{barry2013manipulation, king2016rearrangement} or bias action sampling near objects~\cite{huang2019large}.
A more intuitive \emph{object-centric} paradigm~\cite{saxena2023planning}, 
inspired by how humans reason about rearrangement, 
decouples the problem into two steps: 
first, it searches for desired object motions by assuming objects can move on their own (i.e., without considering the robot); 
then, it computes the robot motions needed to realize those object trajectories.
This paradigm is most methodologically aligned with our proposed framework.
However, unlike~\cite{saxena2023planning}, which was open-loop and originally developed for object retrieval in confined spaces, 
our framework addresses a broader range of large-scale rearrangement tasks involving intensive object-object interactions and real-world uncertainties, by employing closed-loop action generation in real time.

\begin{figure*}[!hbt]
\centering
\small
\begin{center}
\begin{tabular}{ c || c | c | c |  c |  c |  c | c | c}
\hline
\multirow{2}{*}{} & \multirow{2}{*}{Object-Centric} & \multirow{2}{*}{Nonprehensile} & \multirow{2}{*}{Data-Driven} & \multirow{2}{*}{Closed-Loop} & \multirow{2}{*}{Versatile} & \multirow{2}{*}{Explicit-Goal} & \multicolumn{2}{c}{Max $\#$ Objects}\\
\cline{8-9}
 & & & & & & & Sim & Real\\
\hline
~\cite{wang2021uniform, gao2023minimizing} & \cmark & \xmark & \xmark & \xmark & \xmark & \cmark & 30 - 50 & 12 \\
~\cite{haustein2015kinodynamic, king2015nonprehensile} & \xmark & \cmark & \xmark & \xmark & \xmark & \xmark & 6 - 7 & 7 \\
~\cite{king2016rearrangement} & \cmark & \cmark & \xmark & \xmark & \xmark & \xmark & 7 & 1 \\
~\cite{moll2017randomized, agboh2018real} & \xmark & \cmark & \xmark & \cmark & \xmark & \xmark & 15 - 40 & 9 - 16 \\
~\cite{zeng2018learning, eitel2020learning, song2020multi, huang2022interleaving} & \xmark & \cmark & \cmark & \cmark & \xmark & \xmark & 11 - 40 & 8 - 30 \\
~\cite{huang2019large} & \cmark & \cmark & \xmark & \xmark & \cmark & \cmark & 100 & 32 \\
~\cite{pan2021decision} & \xmark & \cmark & \xmark & \cmark & \xmark & \xmark & 200 & 20 \\
~\cite{vieira2022persistent} & \xmark & \cmark & \xmark & \xmark & \xmark & \xmark & 10 & -- \\
~\cite{ren2022rearrangement, ren2023kinodynamic} & \xmark & \cmark & \xmark & \cmark & \cmark & \xmark & 36 & 20 \\
~\cite{saxena2023planning} & \cmark & \cmark & \xmark & \xmark & \xmark & \xmark & 16 & 5 \\
Ours & \cmark & \cmark & \xmark & \cmark & \cmark & \xmark & 100 & 32 \\
\hline
\end{tabular}
\end{center}
\caption{Qualitative comparison of the state-of-the-art rearrangement solutions. 
Each column represents one characteristic of the solution, from left to right:
1) \emph{Object-Centric:} the proposed method incorporates object-centric components (e.g., sampling strategy, action primitives, planning paradigm, etc), otherwise, robot-centric;
2) \emph{Nonprehensile:} nonprehensile actions are incorporated;
3) \emph{Data-Driven:} the method is data-driven and requires extra time for training;
4) \emph{Closed-Loop:} the method generates closed-loop motion plans that can handle real-world uncertainties;
5) \emph{Versatile:} the method has shown the capability to transfer across different rearrangement tasks (e.g., relocating, separating, sorting, etc);
6) \emph{Explicit-Goal:} the method requires an explicit goal pose or location for each object.
7) \emph{Max \# Objects:} the maximum number of movable objects the method can deal with, as shown by the corresponding (simulation and real-world) experiments.}
\label{tab:related_work}
\end{figure*}

While various nonprehensile rearrangement planners have been investigated, 
most of them are limited in problem scale or robustness against uncertainties.
Some are tailored to specific, small-scale tasks (such as retrieving~\cite{dogar2012physics} or pushing a single target object~\cite{king2015nonprehensile}) and tend to become inefficient in more cluttered and complex scenarios.
Others rely on explicitly defined goal locations to guide sampling, which restricts their flexibility in more general rearrangement problems~\cite{huang2019large}.
Moreover, many of these methods are open-loop, making them vulnerable to real-world uncertainties and modeling errors, especially in physics-intensive and longer-horizon tasks where errors can easily accumulate.
To this end, in this work, we propose an object-centric planning (OCP) framework that
\begin{enumerate}
    \item as a unified framework, efficiently addresses a range of planar nonprehensile rearrangement problems (e.g., with or without explicit goals, single or multiple target objects) previously handled by separate approaches;

    \item reacts to perception and physical uncertainties and is robust against modeling inaccuracies, through real-time actions generated in a closed-loop;

    \item generalizes and scales to various large-scale rearrangement tasks with different task objectives.
\end{enumerate}

\emph{Limitations and Generalizability.}
The scope of our proposed framework is focused on 2D planar (e.g., tabletop) pushing-based nonprehensile rearrangement problems. 
While incorporating other types of motions (such as sliding or pivoting) could extend its applicability to more general object rearrangement scenarios, 
those actions are beyond the scope of the current work. 
Nonetheless, within this category, 
our framework is highly generalizable: 
it can be applied to a rearrangement problem as long as the task objective can be described by a heuristic function (as will be introduced in Sec.~\ref{sec:obj_centric}), 
including various settings where one, some, or all objects are considered target objects and whether or not explicit goal poses are specified (see example tasks in Sec.~\ref{sec:tasks}).

The rest of the paper is structured as follows. 
We review related literature in Sec.~\ref{sec:relatedwork} and provide preliminaries in Sec.~\ref{sec:prelim}.
Under our \emph{object-centric} paradigm, we formulate the nonprehensile rearrangement planning problem in Sec.~\ref{sec:prob}.
We present the algorithmic details of our proposed planner in Sec.~\ref{sec:method}, and introduce how real-time robot actions are generated for execution in Sec.~\ref{sec:action}.
By experimental comparisons with state-of-the-art methods in simulation and real world, we evaluate our planner in Sec.~\ref{sec:experiemnts}.
Furthermore, based on our highly efficient rearrangement solution and extensive experimental results, we propose a benchmarking protocol in Sec.~\ref{sec:benchmark} to facilitate research in nonprehensile rearrangement.
Finally, we conclude in Sec.~\ref{sec:conclusion}.

\section{Related Work}
\label{sec:relatedwork}

The table in Fig.~\ref{tab:related_work} qualitatively compares state-of-the-art rearrangement solutions with our proposed planner across several key aspects, as shown in the table columns.
Our proposed planner is the only large-scale nonprehensile solution that can generalize to different task setups without necessarily requiring explicit goal definition, while not being data-consuming.
Next, we review related literature from multiple perspectives.

\emph{Nonprehensile Manipulation and Planar Pushing:}
Prehensile manipulation, which relies on robotic grasping, has been extensively studied to facilitate many manipulation tasks~\cite{gao2022fast, wang2022efficient, wen2022catgrasp}.
As complementary skills to prehensile manipulation, 
nonprehensile manipulation (defined as manipulation without grasping~\cite{mason1999progress}) such as pushing~\cite{dengler2022learning}, sliding~\cite{hang2019pre, song2020probabilistic, yi2023precise}, and pivoting~\cite{hou2020robust, zhang2023learning},
has also been actively explored, to expand the possibilities from single object reconfiguration~\cite{doshi2022manipulation} to large-scale object rearrangement in clutters~\cite{papallas2020non, paus2020predicting, saleem2020planning, lee2021tree}.
Among them, 
pushing has gained particular attention for its capability to operate in confined workspaces~\cite{shome2019towards}.
Planar pushing of a single object is one of the simplest scenarios of pushing-based manipulation.
Existing works have developed analytical models~\cite{ruggiero2018nonprehensile, zhou2018convex, zhou2019pushing, chavan2020planar, halm2020quasi, bertoncelli2020linear, hogan2020reactive} to address the precise planar pushing problem.
However, since they rely on known contact geometry and physical properties, and are typically derived under over-simplified assumptions (e.g., convex shapes),
these analytical approaches do not scale well to real-world scenarios.
More recently, data-driven methods~\cite{yu2016more, bauza2018data, ajay2018augmenting, bauza2019omnipush, kloss2022combining} have been studied for planar pushing under more realistic challenges.
However, the learned models in general require vast amounts of data for training and are hard to generalize over different task setups and perception domains without substantial fine-tuning.
So far, the research on planar pushing mainly focuses on applications involving a single target object, 
while the developed pushing strategies have not been widely extended to large-scale problems such as rearrangement to concurrently manipulate multiple movable objects.

\emph{Kinodynamic Planning:}
When system dynamics is involved due to physical interactions between the robot and its environment, 
modeling of such physics ~\cite{zickler2009efficient, zito2012two} needs to be incorporated into the planning process to guide the motion of the robot.
As such, kinodynamic planning has been proposed to generate actions that comply with the physical constraints of the system~\cite{lavalle2001randomized, bekris2007greedy, lau2009kinodynamic}.
In nonprehensile manipulation problems such as object pushing, however,
manipulation outcomes are often difficult to precisely predict, due to inaccuracies in interaction physics modeling and uncertainties in perception and execution.
Such real-world uncertainties can easily cause the robot motions generated by kinodynamic planning to fail the real-world manipulation.
To address such challenges, some approaches~\cite{koval2015robust, johnson2016convergent, moll2017randomized} incorporate an uncertainty model into the planning process, to generate more conservative motions and reduce the probabilities of execution failure.
Another line of work closes the loop of planning and execution by iterative replanning with receding horizons~\cite{bejjani2018planning, ren2022rearrangement, ren2023kinodynamic}.

\emph{Rearrangement-based Manipulation:}
Object rearrangement is common and important,
involving manipulation of small objects in a confined space~\cite{wang2022lazy, wada2022reorientbot} or heavy objects in a large space~\cite{szot2021habitat, gan2022threedworld}.
When relying solely on prehensile actions (e.g., pick-and-place) and without considering the physics, 
the rearrangement problem can be reduced to a geometric problem with a discrete action space, and long-horizon problems can be solved efficiently under such simplifications. 
Some prehensile approaches use graph-based or tree-based search~\cite{han2018complexity, labbe2020monte, nam2021fast, shome2021fast, wang2021uniform, tian2022sampling, gao2023minimizing, halperin2023space} to provide near-optimal solutions;
some others use learned models~\cite{zeng2021transporter, qureshi2021nerp, danielczuk2021object, liang2022learning, zhang2022visually, ding2023task} to handle more complex contraints.
However, pick-and-place type of prehensile actions can be infeasible when objects are not graspable by the robot, e.g., in a tightly packed environment with limited free space.
By incorporating nonprehensile actions such as pushing, 
more diverse and efficient solutions can be generated.
Using approximated analytical models or physics engines to predict the outcomes of physical interactions,
sampling-based planners~\cite{haustein2015kinodynamic, king2015nonprehensile, vieira2022persistent} have been proposed to generate open-loop motion plans.
However, due to inaccuracies in physics modeling and real-world uncertainties, 
the generated plans often fail in execution,
especially in long-horizon tasks where errors significantly accumulates.
To improve robustness against real-world uncertainties through closed-loop execution, 
strategies such as online replanning~\cite{agboh2018real}, 
distance-guided greedy local search~\cite{huang2019large, pan2021decision},
interleaving progress-controlled planning and execution~\cite{ren2022rearrangement, ren2023kinodynamic} have been proposed.
Driven by the recent advance in deep learning, nonprehensile rearrangement policies can be learned from demonstration or experience data to facilitate tasks such as pushing-based object relocation~\cite{yuan2019end}, multi-object sorting~\cite{song2020multi, han2021toward, tang2023selective}, object singulation for retrieval~\cite{eitel2020learning, huang2022interleaving} and object separation for clutter removal~\cite{zeng2018learning, huang2021dipn}.
These data-driven approaches enable real-time action generation directly from raw image inputs.
However, they are generally data-consuming and difficult to transfer in different task setups~\cite{kaelbling2020foundation}.
For example, a policy trained for separating objects can be inefficient in relocating an object.

\section{Preliminaries}
\label{sec:prelim}

This work aims to address the problem of nonprehensile pushing-based object rearrangement on a 2D plane (e.g., tabletop), 
where a robot manipulator interacts with $N$ movable objects in a bounded workspace $\mathcal{W} \subset \mathbb{R}^2$.
We assume the objects are moved by the robot in a quasi-static manner without rolling or flipping.
The objective of the problem is to find a sequence of feasible robot actions 
that rearranges the objects into desired states while keeping them within $\mathcal{W}$.

\subsection{Kinodynamic Planning for Nonprehensile Rearrangement}
\label{sec:repre}

The nonprehensile rearrangement problem is conventionally formulated as a kinodynamic planning problem,
where higher-order derivatives (e.g., velocities of the robot and objects) need to be included in the state space to account for system dynamics.
However, following the quasi-static assumption and similar to~\cite{king2015nonprehensile, king2016rearrangement}, 
we simplify the problem by reducing the state space to configurations (of the robot and objects) only, excluding their velocities.
As such, the simplified kinodynamic planning problem is defined on the composite configuration space of the robot and all objects, 
for which we introduce the following definitions:

\subsubsection{Configuration Space of the Robot}
Formally, the robot's configuration space is denoted by $\mathcal{Q}^R \subset \mathbb{R}^M$ where $M \in \mathbb{Z}^+$ is the robot's degrees of freedom.
$\mathcal{Q}^R_{free} \subset \mathcal{Q}^R$ is the free configuration space of the robot in which the robot does not collide with itself or the static environment, nor exceed its joint limits.
Note that contacts between the robot and movable objects are allowed as they are necessary for manipulation.
The robot state at time $t$ is denoted by $\bm{q}_t \in \mathcal{Q}^R$.

\subsubsection{Configuration Space of All Objects}
For a single object, we denote its configuration space by $\mathcal{Q}^i \in SE(2)$ where $i = 1, \cdots, N$.
The composite configuration space of all movable objects is then represented by the Cartesian product $\mathcal{Q}^O = \mathcal{Q}^1 \times \cdots \times \mathcal{Q}^N$.
The state of all objects at time $t$, called an arrangement, is denoted by $\bm{s}_t = \left(\bm{s}^1_t, \cdots, \bm{s}^i_t, \cdots, \bm{s}^N_t \right) \in \mathcal{Q}^O$, 
where the state $\bm{s}^i_t = \left(\bm{p}^i_t, \theta^i_t \right) \in SE(2)$ of object $i$ includes its position $\bm{p}^i_t = \left(x^i_t, y^i_t\right) \in \mathbb{R}^2$ and orientation $\theta^i_t \in SO(2)$.

The problem space of kinodynamic planning is the composite space $\mathcal{Q} = \mathcal{Q}^R \times \mathcal{Q}^O$ of the entire system, consisting of the configurations of the robot and all objects.
A system state at time $t$ is valid if the robot is collision-free $\bm{q}_t \in \mathcal{Q}^R_{free}$ and all object positions remain within the bounded workspace $\bm{p}^i_t \in \mathcal{W}$, $\forall i$.
Collisions between objects are allowed to enable concurrent object-object interactions.

\subsubsection{Action Space}
The action space $\mathcal{A}$ consists of all actions the robot is allowed to perform, 
which can be represented differently depending on the task-specific action definition.
For an $M$-DoF robot manipulator, a common choice is the joint velocity space, i.e., $\mathcal{A} \subset \mathbb{R}^M$, 
where an action $\bm{a}_t \in \mathcal{A}$ is an instantaneous joint velocity commanded to move the robot.

\subsubsection{System Dynamics}
Under the quasi-static assumption, the entire system composing the robot and all movable objects is modeled as a discrete-time dynamic system,
represented by a transition function $\Gamma: \mathcal{Q}^R \times \mathcal{Q}^O \times \mathcal{A} \mapsto \mathcal{Q}^R \times \mathcal{Q}^O$ in Eq.~\eqref{eq:robot_trans}:
\begin{equation}
\label{eq:robot_trans}
    \left(\bm{q}_{t+1}, \bm{s}_{t+1}\right) = \Gamma\left(\bm{q}_t, \bm{s}_t, \bm{a}_t\right)
\end{equation}
which infers the outcome robot state $\bm{q}_{t+1}$ and arrangement $\bm{s}_{t+1}$ (i.e., the outcome state of all objects), given the current robot state $\bm{q}_t$, the current arrangement $\bm{s}_t$, and action $\bm{a}_t$ executed by the robot, according to real-world physics laws.

\subsubsection{Goal Criterion}
To assess whether all objects are successfully rearranged,
we define a criterion function $g: \mathcal{Q}^O \mapsto \{0, 1\}$,
where $g(\bm{s}_t) = 1$ indicates that an arrangement $\bm{s}_t \in \mathcal{Q}^O$ satisfies the task goal.
Some prior works instead specify an explicit goal region $\mathcal{G}_i \subset \mathcal{W}$ for each object $i$ and check whether the object lies within it to determine the task completion~\cite{king2015nonprehensile, huang2019large}.
However, 
for other tasks, such as those requiring only relative configurations of some objects~\cite{song2020multi},
explicitly defining $\mathcal{G}_i$ is unnecessary and difficult (see example tasks in Sec.~\ref{sec:experiemnts}).
Hence,
a criterion function $g$ offers a more general and flexible representation of rearrangement goals.

We consider a robot starting at its initial configuration $\bm{q}_0$, 
tasked with rearranging multiple objects from an initial arrangement $\bm{s}_0 = \left(\bm{s}_0^1, \cdots, \bm{s}_0^i, \cdots, \bm{s}_0^N\right)$.
The goal of nonprehensile rearrangement problem can be formally defined as finding a sequence of $T$ robot actions, denoted as $\tau = \{\bm{a}_0, \cdots, \bm{a}_t, \cdots \bm{a}_{T-1}\}$, such that:
\begin{enumerate}
    \item The system state transitions under the system dynamics by $\left(\bm{q}_{t+1}, \bm{s}_{t+1}\right) = \Gamma\left(\bm{q}_t, \bm{s}_t, \bm{a}_t\right)$, $t = 0, \cdots, T-1$.
    \item For all intermediate time steps $t = 1, \cdots, T$, the system state is always valid: The robot is in its free C-space $\bm{q}_t \in \mathcal{Q}^R_{free}$, and all objects are within the workspace $\bm{p}^i_t \in \mathcal{W}$, $\forall i = 1, \cdots, N$.
    \item The final arrangement, after the sequence of robot actions $\tau$ has been executed in the desired order, satisfies the goal criterion, i.e., $g(\bm{s}_T) = 1$.
\end{enumerate}

\subsection{Nonprehensile Object Pushing}
\label{sec:uno_push}

\begin{figure}[t]
    \centering
    \includegraphics[width=\columnwidth]{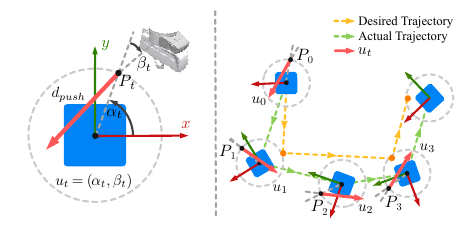}
    \caption{\emph{Left:} A robot pushing action is represented by $\bm{u}_t = \left(\alpha_t, \beta_t\right)$, where $\alpha_t$ and $\beta_t$ are two angles specified in the object's body frame. 
    To perform $\bm{u}_t$, the robot needs to first place its pusher at the position $P_t$ (determined by $\alpha_t$) and then translate in the direction of $\beta_t$ by a distance $d_{push}$ to interact with the object (blue cube) through the push (orange arrow);
    \emph{Right:} By consecutively generating and executing the pushing actions $\bm{u}_0$ through $\bm{u}_3$, the robot can push the object to follow a desired reference trajectory (yellow dashed lines), as a sequence of desired poses or positions of the object.
    With a closed-loop pushing strategy, the object's actual trajectory (green dashed lines) due to execution will not deviate much from the reference.}
    \label{fig:uno_push}
\end{figure}

In general, consider a single object currently at a state $\bm{s}_t \in SE(2)$ and given a desired state $\hat{\bm{s}}_{t+1} \in SE(2)$ for the next time step,
the desired motion of the object can be accordingly calculated by $\Delta \bm{s}_t = \left(\bm{s}_t \right)^{-1} \cdot \hat{\bm{s}}_{t+1} \in SE(2)$, as a relative transformation expressed in the object's body frame.
With an off-the-shelf pushing strategy, the desired pushing action $\bm{u}_t$ should be generated based on the desired $\Delta \bm{s}_t$.
We generally denote this process by a function $\bm{u}_t \gets \Call{PushStrategy}{\bm{s}_t, \hat{\bm{s}}_{t+1}}$.
As shown in Fig.~\ref{fig:uno_push} (left), 
the robot's pushing action can be parameterized by two angles defined in the object's body frame, $\bm{u}_t = (\alpha_t, \beta_t)$.
The first angle $\alpha_t$ determines the robot pusher's starting position relative to the object, i.e., a point $P_t \in \mathbb{R}^2$ on a circle centered around the object, where the circle's radius corresponds to the object's size;
the second angle $\beta_t$ specifies the pushing direction.
To execute $\bm{u}_t$,
the robot needs to first place its pusher (or end-effector) to the starting position $P_t$ (computed based on $\alpha_t$),
and then move its pusher in the pushing direction $\beta_t$ for a fixed distance $d_{push}$ along the workspace plane.
The pushing distance $d_{push}$ is a preset hyperparameter;
given a fixed pushing frequency, a larger $d_{push}$ is typically used when faster pushing is desired, 
while a smaller value is preferred when higher pushing precision is required.
Note that $\bm{u}_t$ differs from the lower-level robot action $\bm{a}_t$ (e.g., joint velocities) defined in Sec.~\ref {sec:repre}. 
A robot controller (e.g., a Cartesian position controller) will convert $\bm{u}_t$ to $\bm{a}_t$, which can be directly commanded to the robot for execution. 

By iteratively calling the pushing strategy and executing the generated actions in a closed loop,
the robot can push the object to follow a desired trajectory.
As shown in Fig.~\ref{fig:uno_push} (right), the object's pose is tracked in real-time by sensors such as cameras.
Given a desired trajectory of the object, represented by an ordered sequence of desired waypoints,
the robot is able to continuously push the object along this trajectory,
through consecutively adjusting its pusher to the positions $P_0, \cdots, P_3$ and executing the corresponding actions $\bm{u}_0, \cdots, \bm{u}_3$ in between.
In this work, we use the UNO Push framework from our previous work~\cite{wang2024uno} as the pushing strategy.

\section{Problem Statement}
\label{sec:prob}

Inspired by human strategies in rearranging objects,
we propose to adopt an \emph{object-centric} planning paradigm, as formulated in Sec.~\ref{sec:obj_centric},
to address large-scale nonprehensile rearrangement problems. 
Furthermore,
to handle real-world uncertainties,
we close the planning loop by interleaving it with real execution as detailed in Sec.~\ref{sec:interleaved}.

\subsection{Object-Centric Planning}
\label{sec:obj_centric}
Traditional sampling-based approaches are mostly \emph{robot-centric},
where robot actions are sampled based solely on robot-relevant constraints, without explicitly targeting interactions with specific objects~\cite{haustein2015kinodynamic, king2015nonprehensile}.
The motions of surrounding objects, 
resulting from these sampled robot actions, 
are passively predicted by an approximate model of system dynamics $\Gamma$.
As such,
the generated robot actions are not sufficiently guided by how the objects need to be reconfigured;
consequently,
\emph{robot-centric} strategies are often inefficient,
due to their lack of active exploitation of task-relevant information (e.g., the desired object reconfigurations).

In this work, we approach the nonprehensile rearrangement problem using an \emph{object-centric} formulation.
In short, 
we first search to figure out how the objects need to be moved to accomplish the rearrangement task, 
by planning their desired trajectories independently of the robot.
Then, using a nonprehensile object pushing strategy (e.g., the UNO Push framework~\cite{wang2024uno}),
the robot generates and executes actions in a closed-loop manner, to move each object along its desired trajectories planned beforehand.

To plan objects' desired trajectories, we assume each object can move autonomously on a flat surface (i.e., without needing to be moved by a robot),
as if we equip each object with a virtual actuator to enable its translation and rotation in all directions.
The desired trajectory of an object is represented by an ordered sequence of multiple waypoints in $SE(2)$, denoted by $\mathcal{T}^i = \{\hat{\bm{s}}^i_1, \cdots, \hat{\bm{s}}^i_k, \cdots, \hat{\bm{s}}^i_K\}$.
Each waypoint $\hat{\bm{s}}^i_k = \left(\hat{\bm{p}}^i_k, \hat{\theta}^i_k \right) = \left(\hat{x}^i_k, \hat{y}^i_k, \hat{\theta}^i_k \right) \in SE(2)$ is a desired intermediate configuration of the object along the trajectory.
If under perfect control and execution, the object is expected to reach each waypoint one by one from $\hat{\bm{s}}^i_1$ to $\hat{\bm{s}}^i_K$.

We assume only one object is activated at a time to actively move.
The activated object does not have to be collision-free while following its desired trajectory since collisions between objects are allowed as aforementioned.
As such,
other objects may be passively moved due to object-object interactions.
We represent such dynamics involving only mutual interactions between objects by an object-centric transition function $\Pi : \mathcal{Q}^O \times \{1, \cdots, N\} \times \Xi \mapsto \mathcal{Q}^O$, where $\Xi$ is the space of all possible object trajectories:
\begin{equation}
    \bm{s}_{t+1} = \Pi \left(\bm{s}_t, i, \mathcal{T}^i \right)
\label{eq:obj_centric_trans}
\end{equation}
which infers the outcome arrangement $\bm{s}_{t+1} \in \mathcal{Q}^O$, given the current arrangement $\bm{s}_t \in \mathcal{Q}^O$, the index $i$ of the activated object, and its trajectory $\mathcal{T}^i$.
We model $\Pi$ by a physics engine, more details of which will be given in Sec.~\ref{sec:experiemnts}.
It is worth noting that, different from the robot-centric transition function $\Gamma$ defined in Eq.~\eqref{eq:robot_trans}, 
this object-centric transition function $\Pi$ does not model any interactions involving the robot.
This modeling simplification in $\Pi$ introduces more inaccuracy compared to real-world physics.
However, by interleaving planning and execution (as described in Sec.~\ref{sec:interleaved}),
the errors due to such modeling inaccuracy can be adequately mitigated.

\subsection{Interleaved Planning and Execution}
\label{sec:interleaved}

For long-horizon and physics-intensive manipulation tasks (such as rearranging a large number of objects), the execution will accumulate errors that deviate the real-world system transitions from what is predicted by the planner,
due to inaccurate physics modeling and imperfect perception.
As a result, the robot will likely fail the task even if a solution has been found.
To address this,
rather than executing the entire plan at once,
we interleave planning and real execution to progressively guide the system towards the task goal.
As such,
the planning horizon before each execution has to be limited, 
and a heuristic function to quantitatively evaluate and monitor the planning progress is needed.
We denote the task-dependent heuristic function by $h: \mathcal{Q}^O \mapsto \mathbb{R}$, which evaluates a cost given an arrangement $\bm{s}_t \in \mathcal{Q}^O$.
A smaller value of $h\left(\bm{s}_t \right)$ indicates that the system state gets closer to achieving the task goal.
As will be shown later in Sec~\ref{sec:experiemnts},
some simple yet effective distance-based functions can be easily used as heuristics for various practical rearrangement problems.

In each planning cycle, we adopt a greedy objective by minimizing the heuristic cost $h$ while respecting system physics and constraints.
Mathematically, this can be represented as a constrained optimization formalization in Eq.~\eqref{eq:optimize}.
\begin{subequations}
\label{eq:optimize}
\begin{align}
    \underset{{\{\left(i_d, \mathcal{T}^{i_d} \right)\}_{d=1}^D}}{\textrm{minimize}} \quad & h(\bm{s}_{t+D}) \label{eq:optimize_a}\\
    \textrm{subject to} \quad & 0 \leq D \leq D_{\max}, \label{eq:optimize_b}\\
                  \quad & \forall d = 1, \cdots, D: \label{eq:optimize_c} \\
                  \quad & \quad \quad \bm{s}_{t+d} = \Pi\left(\bm{s}_{t+d-1}, i_d, \mathcal{T}^{i_d} \right), \label{eq:optimize_d} \\ 
                  \quad & \quad \quad \bm{p}_{t+d}^i \in \mathcal{W}, \quad \forall i \in \{1, \cdots, N\} \label{eq:optimize_e}
\end{align}
\end{subequations}
Specifically, given the current arrangement $\bm{s}_t$, the optimization problem aims to find an ordered sequence of $D$ object-trajectory pairs, i.e., $\{\left(i_d, \mathcal{T}^{i_d} \right)\}_{d=1}^D$,
such that the heuristic cost $h(\bm{s}_{t+D})$ will be minimized after these $D$ trajectories have been performed.
Each pair in this sequence,
$(i_d, \mathcal{T}^{i_d})$, consists of the index $i_d \in \{1, \cdots, N\}$ of the activated object and a desired trajectory $\mathcal{T}^{i_d}$ planned for this object.
The length of the sequence, denoted by an integer variable $D$, is the horizon of the planning cycle,
which is equal to the number of decisions made by the robot to switch to manipulating a different target object.
We limit the value of $D$ not to exceed a threshold $D_{\max}$ in Eq.~\eqref{eq:optimize_b}, 
to ensure the planning horizon is not too long; 
otherwise, the real execution can accumulate large errors and greatly deviate the system from its planned manipulation results,
leading to inefficient or even failed real-world rearrangement.
Eq.~\eqref{eq:optimize_d} constrains the system to transit under $\Pi$ to comply with the physics laws,
and Eq.~\eqref{eq:optimize_e} confines each object to be within the robot's workspace.

\begin{figure}[t]
\begin{algorithm}[H]
\small
\caption{Object-Centric Rearrangement Planning}
    \begin{algorithmic}[1]
        \Require Initial arrangement $\bm{s}_0$, goal criterion $g(\cdot)$
        \Ensure Task completion ($true$ or $false$)
        \State $\bm{s}^* \gets \bm{s}_0$
        \While{$\Call{Time.Available}{\null}$ and $g(\bm{s}^*) == 0$}
        
        \hfill \Comment{Task not Accomplished}
            \State $\{(i_d, \mathcal{T}^{i_d})\}_{d=1}^D \gets \Call{OCP}{\bm{s}^*}$ \hfill \Comment{Sec.~\ref{sec:method}}
            \For{$d = 1, \cdots, D$}
                \State $\Call{Execute}{\bm{s}^*, i_d, \mathcal{T}^{i_d}}$ \hfill \Comment{Sec.~\ref{sec:action}}
                \State $\bm{s}^* \gets \Call{ObserveObjects}{\null}$ \hfill \Comment{via Real-time Sensing}
            \EndFor
        \EndWhile
    \State \Return $true$ \textbf{if} $g(\bm{s}^*) == 1$ \textbf{else} $false$
    \end{algorithmic}
\label{alg:rearrangement}
\end{algorithm}
\end{figure}

The pipeline for nonprehensile rearrangement planning under our \emph{object-centric} formulation is outlined in Alg.~\ref{alg:rearrangement}.
Our framework starts with all objects being at states in $\bm{s}_0 \in \mathcal{Q}^O$ (i.e., the initial arrangement).
Without considering the robot, our object-centric planner (OCP) will first search for an ordered sequence of $D$ object-trajectory pairs $\{\left(i_d, \mathcal{T}^{i_d} \right)\}_{d=1}^D$ by minimizing the heuristic cost $h$, as will be detailed in Sec.~\ref{sec:method}.
Then the robot will execute by calling the pushing strategy to make each activated object follow their desired trajectory one by one in the order of $d = 1, \cdots, D$, while strictly respecting the robot's physical and kinematic constraints, as will be described in Sec.~\ref{sec:action}.
After each execution, the real-world arrangement $\bm{s}^*$ (i.e., the state of all objects) will be observed by sensors for starting the subsequent planning.
The robot will repeat this procedure of interleaving planning and execution until it accomplishes the rearrangement task or exceeds the time budget.
As such, by sensing the real-world state and adaptively replanning and adjusting its actions,
the robot reacts to discrepancies between the planned motions and real execution and continuously eliminates accumulated errors in the recent execution.

\section{Object-Centric Sampling-based Planner}
\label{sec:method}

As formulated in Sec.~\ref{sec:prob},
we need to plan the desired object trajectories without considering the robot,
by minimizing the heuristic cost as presented in Eq.~\eqref{eq:optimize}.
In practice, finding the optimal solution is intractable and usually not needed.
As long as the heuristics cost $h$ is progressively minimized, the rearrangement task can be solved efficiently.
We propose to use a sampling-based approach to search the desired object trajectories as a solution to Eq.~\eqref{eq:optimize}.

\begin{figure*}[t]
    \centering
    \includegraphics[width=1.6\columnwidth]{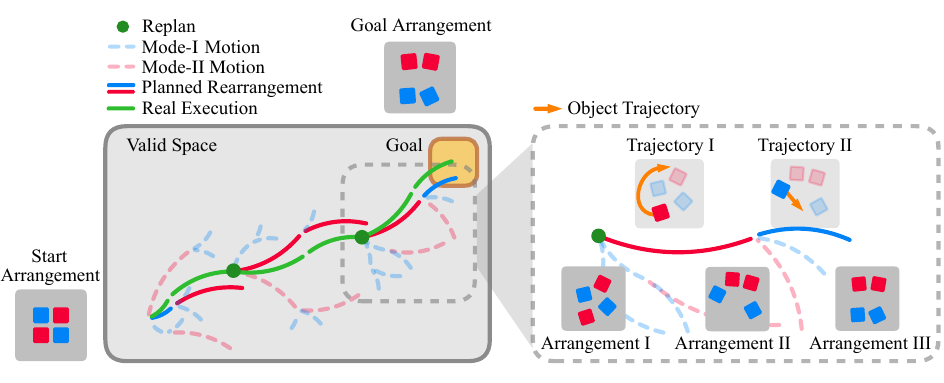}
    \caption{A schematic plot of OCP.
    In the left figure, a motion tree is progressively grown from the start arrangement (lower left) towards the goal arrangement (upper right).
    Each edge of the tree is an explored object motion (i.e., trajectory) generated by one of two exploration modes, \emph{Mode I} (blue line) and \emph{Mode II} (red line).
    Each edge leads to an outcome arrangement represented by a tree node.
    Replanning (green dots) by sensing real-world arrangement is needed to eliminate the errors between the planned rearrangement solution (solid blue and red lines) and real execution (green lines) due to real-world uncertainties.
    The right figure shows the planned rearrangement since the last replanning, which consists of two consecutive object motions.
    The first (solid red line) is an \emph{Mode II} motion that moves the red cube through a curvy Trajectory I (orange), resulting in an arrangement shown by Arrangement II;
    the second motion (solid blue line) is under \emph{Mode I}, and leads to the arrangement shown in Arrangement III by moving the blue cube through a straight-line Trajectory II (orange).}
    \label{fig:tree}
\end{figure*}

Specifically, we develop an object-centric sampling-based planner (OCP) by maintaining a tree-based data structure denoted by $Tr$.
Each node of the tree, denoted by $n \in Tr.\Call{Nodes}{\null}$, represents an explored arrangement.
The arrangement associated with a node can be accessed through $n.\bm{s} \in \mathcal{Q}^O$.
Each edge in the tree directs from one node to its child node, representing an object-trajectory pair $\left(i_d, \mathcal{T}^{i_d}\right)$ that transits the system (i.e., all objects) to the arrangement associated with the child node.
The algorithmic steps of OCP are outlined in Alg.~\ref{alg:OCP}, and a schematic plot is shown in Fig.~\ref{fig:tree}.

\begin{figure}[t]
\begin{algorithm}[H]
\small
\caption{Object-Centric Planner -- OCP($\cdot$)}
    \begin{algorithmic}[1]
        \Require The current arrangement $\bm{s}^* \in \mathcal{Q}^O$ observed by the sensors
        \Ensure The sequence of the desired object activations and the corresponding trajectories $\{\left(i_1, \mathcal{T}^{i_1}\right), \cdots, \left(i_D, \mathcal{T}^{i_D}\right)\}$
        \State $Tr \gets \{n_{root}.\bm{s} = \bm{s}^*\}$ \hfill \Comment{Add Root Node to Tree}
        \While{$T.\Call{GetSize}{\null} < S_{\max}$}
            \State $n \gets \Call{SampleNode}{Tr}$ \hfill \Comment{Sec.~\ref{sec:node_sampling}}
            \State $i \gets \Call{ActivateObject}{n}$ \Comment{Sec.~\ref{sec:obj_sampling}}
            \State $\bm{s}_o, \mathcal{T}^i \gets \Call{ExpandTree}{n, i}$ \hfill \Comment{Alg.~\ref{alg:expand_tree}}
            \If{$\forall i, \bm{s}_o^i \in \mathcal{W}$ and $\mathcal{T}^i \neq null$}
                \State $n_{new} \gets \Call{CreateNode}{\null}$
                \State $Tr.\Call{AddNode}{n_{new}.\bm{s} = \bm{s}_o}$ \Comment{Add as a New Node}
                \State $Tr.\Call{AddEdge}{(n, n_{new}), \left(i, \mathcal{T}^i \right)}$
            \EndIf
            \If{$g(\bm{s}_o) == 1$} \hfill \Comment{Goal Reached}
                \State $\{\left(i_d, \mathcal{T}^{i_d} \right)\}_{d=1}^D \gets Tr.\Call{BackTrace}{n_{new}}$
                \State \Return $\{\left(i_d, \mathcal{T}^{i_d} \right)\}_{d=1}^D$
            \EndIf
        \EndWhile
        \State $n^* \gets \arg\min_{n \in Tr.\Call{Nodes}{\null}} h(n.\bm{s})$
        \State $\{\left(i_d, \mathcal{T}^{i_d}\right)\}_{d=1}^D \gets Tr.\Call{BackTrace}{n^*}$
        \State \Return $\{\left(i_d, \mathcal{T}^{i_d} \right)\}_{d=1}^D$
    \end{algorithmic}
\label{alg:OCP}
\end{algorithm}
\end{figure}

The tree is initialized with a root node $n_{root}$ corresponding to the current start arrangement $\bm{s} \in \mathcal{Q}^O$ observed by sensors.
We limit the tree size (i.e., the number of tree nodes) by $S_{\max}$.
While the tree size has not reached the limit $S_{\max}$, OCP keeps exploring through the following major steps:
First, a node $n$ will be randomly sampled to grow the tree for exploration, as introduced in Sec.~\ref{sec:node_sampling};
Since only one object is activated each time to actively move, one object $i \in \left\{1, \cdots, N \right\}$ will be selected through sampling in Sec.~\ref{sec:obj_sampling};
Next, from the sampled node, a trajectory $\mathcal{T}^i$ of the activated object will be simulated with two possible exploration modes in Sec.~\ref{sec:tree_expand}.
As will be detailed in Sec.~\ref{sec:a_star}, one of the exploration modes empowered by a soft-$A^*$ algorithm, is an important design and crucial to the effectiveness of the planner.
The outcome arrangement $\bm{s}_o \in \mathcal{Q}^O$ after simulating the trajectory, if valid (i.e., all objects are within the workspace $\mathcal{W}$), will be added as a new node into the tree.
If the arrangement $\bm{s}_o$ of the newly added node satisfies the goal criterion, the tree expansion will be stopped and we will backtrace this new node to extract the solution;
otherwise, this tree expansion procedure will be repeated until the tree size reaches $S_{\max}$.
Finally, we will backtrace from the tree node that has the smallest heuristic cost, denoted by $n^*$,  to extract the planned object trajectories $\{\left(i_d, \mathcal{T}^{i_d} \right)\}_{d=1}^D$ as the solution to Eq.~\eqref{eq:optimize}.

\subsection{Node Sampling for Tree Expansion}
\label{sec:node_sampling}
For each node $n$ of the tree, we use $n.D$ to denote the depth of the node $n$ in the tree, and $n.N_c$ to denote the number of the child nodes of the node $n$.
Similar to Expansive Space Tree (EST)~\cite{hsu1997path}, we associate a weight $w(n)$ for each node in Eq.~\eqref{eq:node_weight}.
The probability of sampling a node is equal to its normalized weight, i.e., $P(n) = w(n) / \sum_{n' \in Tr.\Call{Nodes}{\null}} w(n')$.
\begin{equation}
\label{eq:node_weight}
w(n) = 
\begin{cases}
    \frac{1}{n.N_c + 1} & n.D < D_{\max}, \\
    0 & \text{otherwise}
\end{cases}\\
\end{equation}
The weight of a node is set to zero if its depth reaches $D_{\max}$, to eliminate this node from being sampled for expansion.
This mechanism limits the tree depth by $D_{\max}$ to be consistent with the limited planning horizon in Eq.~\eqref{eq:optimize}.
Otherwise, the weight is set to be inverse to $n.N_c$.
A large $n.N_c$ indicates that this node has been explored sufficiently with many child nodes, therefore, the weight of this node will be set low so other nodes will have a higher probability of being sampled for exploration.
For nodes that have not been explored so far, they will have zero child nodes.
We add $1$ to $n.N_c$ in the denominator to avoid the issue of dividing by zero.
The weights of nodes will be dynamically updated whenever a new node is added into the tree.

\subsection{Heuristics-Guided Object Activation}
\label{sec:obj_sampling}
To expand from the sampled node $n$ in Sec.~\ref{sec:node_sampling}, the tree will simulate an object trajectory and predict the outcome arrangement $\bm{s}_o$ based on the object-centric transition function $\Pi$.
As aforementioned, only one object can be activated at a time, therefore, an effective sampling policy is needed to select which object to activate.

Given the arrangement $\bm{s}$ associated with the sampled tree node $n$ (i.e., $\bm{s}$ is used to denote $n.\bm{s}$ for simplicity),
we guide the selection of the activated object by the informative gradients of the heuristic function.
In practice, the gradients of the heuristic function can be either analytically derived or numerically approximated via finite differences applied to each dimension of $\bm{s}$;
In our implementation, we use PyTorch's automatic differentiation tool~\cite{paszke2017automatic} to compute these gradients efficiently.
Intuitively, if the gradient magnitude of the heuristic function with respect to the state of the $i$-th object is large, i.e., a large $\lvert \nabla_{\bm{s}^i} h(\bm{s}) \rvert$,
a high probability of sampling this object will be expected since the local change of this object's state can greatly affect the task progress evaluated by the heuristic cost.
In other words, exploiting the motion of objects with high gradient magnitudes is likely to gain fast task progress.
To this end, we integrate the gradients of the heuristic function into a weighted mixture of $N$ Gaussian distributions to model the sampling probabilities for object activation.
Specifically, each Gaussian component corresponds to an object $i$ and is spatially centered at the position of this object. 
The gradient for the $i$-th object, $\lvert \nabla_{\bm{s}^i} h(\bm{s}) \rvert$, is used to model the weight of the $i$-th Gaussian.
The probability $P(i)$ for sampling the $i$-th object to activate is given in Eq.~\eqref{eq:sample_obj_a}.
\begin{subequations}
\label{eq:sample_obj}
\begin{align}
    P(i) &= \frac{1}{Z}\sum_{j=1}^N f\left(\lvert\nabla_{\bm{s}^j} h(\bm{s}) \rvert \right)\cdot \varphi\left(d_{ij}\right) \label{eq:sample_obj_a}\\
    & \propto  f\left(\lvert\nabla_{\bm{s}^i} h(\bm{s}) \rvert \right) + \sum_{j\neq i} f\left(\lvert\nabla_{\bm{s}^j} h(\bm{s}) \rvert \right) \cdot e^{-\frac{d_{ij}^2}{2\sigma^2}} \label{eq:sample_obj_b}
\end{align}
\end{subequations}
where $Z$ is the normalization term to ensure the probabilities for all $i$ sum to $1$; 
$\varphi(\cdot)$ is the density function of a zero-mean Gaussian with variance $\sigma^2$;
$d_{ij}^2 = \lVert \bm{p}^i - \bm{p}^j \rVert^2$ is the squared distance between the object $i$ and $j$;
$f: \mathbb{R}^+ \mapsto \mathbb{R}^{+}$ is a stretching function to exaggerate the magnitude difference between Gaussian components.
In practice, we opt for a simple power function to stretch, i.e., $f(x) = x^k$ $(k > 1)$.

\begin{figure}[t]
    \centering
    \begin{minipage}{0.32\columnwidth}
        \vspace{-40pt}
        \includegraphics[width=\columnwidth]{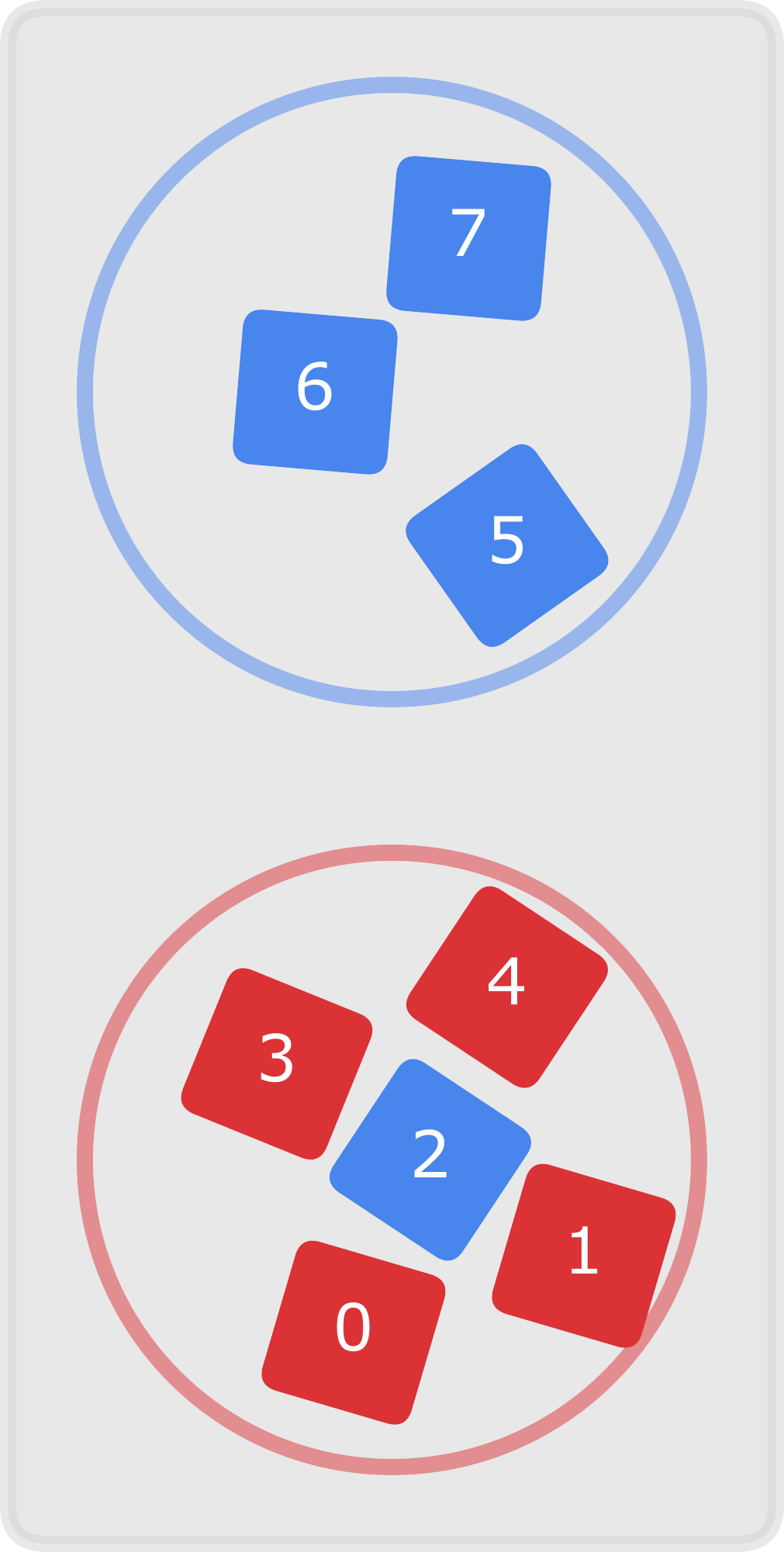}
    \end{minipage}
    \begin{minipage}{0.32\columnwidth}
        \input{figs/plot_obj_pos_2_v.tex}  
    \end{minipage}
    \begin{minipage}{0.32\columnwidth}
        \vspace{-15pt}
        \input{figs/plot_obj_pos_3_v.tex}  
    \end{minipage}
    \vspace{-15pt}
    \caption{
    \emph{Left:} 
    A scenario of the object sorting task.
    Two classes of cubes (red and blue) need to be relocated inside their corresponding goal regions (circles in the same color as the cubes).
    At the current state, all cubes except for Cube $\# 2$ are already sorted.
    \emph{Middle:} The value of $f\left(\lvert\nabla_{\bm{s}^j} h(\bm{s}) \rvert \right)$ for each object, if directly used as the sampling probability, will cause the algorithm trapped by keeping sampling Cube $\# 2$ to activate.
    Since there is no free space around the unsorted Cube $\# 2$ for the robot to approach it, the algorithm will be likely stuck at this point.
    \emph{Right:} Enabled by the weighted mixture of Gaussians, the sampling probabilities of the red cubes ($\# 0$, $\# 1$, $\# 3$, and $\# 4$) surrounding Cube $\# 2$ are increased so that they can be moved to create some free space for Cube $\# 2$, to facilitate the relocation of Cube $\# 2$ to its goal region by the subsequent actions.} 
    \label{fig:prob_obj_case}
\end{figure}

We use Gaussians to model the distribution of Eq.~\eqref{eq:sample_obj} continuously, in order to prevent the sampling from being trapped by a local minimum arrangement.
For example, as showcased in Fig.~\ref{fig:prob_obj_case} (left), the blue cube ($\# 2$) is still far away from its desired goal region (the blue circle on the top) and has a high gradient magnitude, whereas the surrounding red cubes are already placed at their desired locations and thus have low gradient magnitudes.
In the current situation, activating the blue Cube $\# 2$ is ineffective, since there is no free space around it to realize a trajectory achievable by the actual robot.
In this case, directly using the gradient magnitudes $\lvert \nabla_{\bm{s}^i} h(\bm{s}) \rvert$ as the sampling probability is likely to cause the algorithm to get stuck always sampling the ineffective Cube $\# 2$.   
However, smoothened by using Gaussians, the probability of sampling an object depends on not only its own gradient but also the gradients of other objects near it, as can be seen in Eq.~\eqref{eq:sample_obj_b}.
In the above example, using Gaussians can make it possible to activate the surrounding red cubes with high probabilities, 
thus potentially creating some free space around the blue cube for it to be feasibly moved out and toward its desired location.

\begin{figure}[t]
    \centering
    \includegraphics[width=\columnwidth]{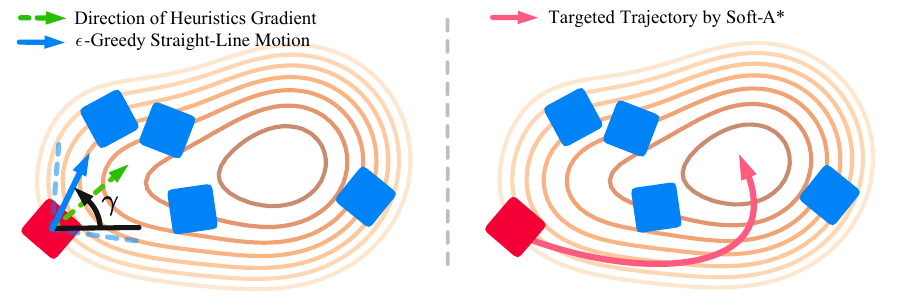}
    \caption{Two exploration modes of generating the object trajectory for expanding the motion tree.
    The contour lines represent the heuristic cost concerning only the activated object (red cube).
    \emph{Left:} With a probability of $1-\epsilon$, \emph{Mode I} samples a straight-line motion (blue solid arrow) with an angle $\gamma$ to the $x$-axis of the workspace (black line).
    The sampling of $\gamma$ is centered around the direction of the heuristic gradient (green dashed arrow) with a range of $\left[-\frac{\pi}{4}, \frac{\pi}{4} \right)$ (between the blue dashed lines);
    \emph{Right:} \emph{Mode II} plans a goal-oriented curvy trajectory (orange curve) for the activated object towards its optimal location (i.e., minimizing the heuristic cost), by using Soft-$A^*$ as will be introduced in Sec.~\ref{sec:a_star}.}
    \label{fig:tree_expand}
\end{figure}

\subsection{Tree Expansion by Analyzing Object Trajectories}
\label{sec:tree_expand}
After an activated object $i$ is selected in Sec.~\ref{sec:obj_sampling}, our planner needs to move this object along a certain trajectory to explore the outcome arrangement.
A straightforward exploration strategy is to move the object locally in a random direction, i.e., a straight-line trajectory.
Such local and random motions of the activated object enable efficient exploration for arrangement by concurrently reconfiguring multiple surrounding objects via object-object interactions.
However, since such explorations are local and the interaction outcomes can be random, it is difficult to find motions that are guaranteed to improve the task progress significantly.

Therefore, we also incorporate a more goal-oriented exploration strategy.
Instead of exploring the local reconfiguration of multiple surrounding objects, the new strategy focuses on the repositioning of the activated object:
Without excessively disturbing other objects, it directly moves the activated object to its desired destination, along an optimal trajectory planned with a certain level of discretization.
Ideally, if each object can be directly moved to its desired location one by one, the rearrangement task can be solved progressively without needing much unnecessary exploration.
Based on this intuition, the second strategy is designed to enable the possibility of finding effective motions that can directly finish the rearrangement of one object (i.e., the activated object).
Fig.~\ref{fig:tree_expand} illustrates the two exploration strategies, and the details are given below:

1) \emph{Mode I: $\epsilon$-Greedy Straight-Line Motion.} 
As illustrated in Fig.~\ref{fig:tree_expand} (left), this mode constructs a straight-line trajectory to guide the motion of the activated object, where the trajectory has only one waypoint $\hat{\bm{s}}^i_1 \in SE(2)$ as the destination.
Rather than purely random,
we adopt a $\epsilon$-greedy policy to determine the direction of the straight line (i.e., an angle $\gamma$) for better exploration efficiency.
With a probability $\epsilon$, $\gamma$ will be uniformly sampled between $[-\pi, \pi)$; with a probability $1 - \epsilon$, $\gamma$ will be greedily sampled around the gradient direction of the heuristic function, i.e., $\left(\frac{\partial h(\bm{s})}{\partial x^i}, \frac{\partial h(\bm{s})}{\partial y^i}\right)$, with an allowed angular deviation between $[-\frac{\pi}{4}, \frac{\pi}{4})$.
The length of the straight line is uniformly sampled from a preset range $[l_{\min}, l_{\max})$, which are hyperparameters chosen according to the size of the workspace and the scale of the problem.
While the activated object translates along with the generated straight line, it also rotates with a constant angular velocity to change its orientation by a randomly sampled $\Delta \theta^i \in [-\pi, \pi)$.

2) \emph{Mode II: Goal-Oriented Trajectory Planned by Soft-$A^*$.} 
Under this mode, the trajectory $\mathcal{T}^i$ is deterministically planned to reach the optimal position of the activated object $i$ while being aware of potential collisions with other objects, as illustrated in Fig.~\ref{fig:tree_expand} (right).
The planning of the trajectory $\mathcal{T}^i$ is solved by a softened version of the $A^*$ algorithm, as will be detailed in Sec.~\ref{sec:a_star}.
Since \emph{Mode II} is deterministic, for the same tree node $n$ and activated object $i$, the planned $\mathcal{T}^i$ is always the same.
To this end, for each tree node $n$, we store the indices of the objects that have been explored under \emph{Mode II} in a set $n.\mathcal{B} \in \{1, \cdots, N\}$.
If a new \emph{Mode II} exploration is going to be performed on an object $i$ that is already explored under \emph{Mode II} (i.e., $i \in n.\mathcal{B}$),
we will skip it to save the computational time cost of a duplicate \emph{Mode II} exploration.

The procedure of generating and simulating a trajectory $\mathcal{T}^i$ for expanding the tree, using one of the two exploration modes, is given in Alg.~\ref{alg:expand_tree}.
Which exploration mode to use is randomly determined, and the probability of using $\emph{Mode II}$ is a hyperparameter $p_{A^*}$.
After $\mathcal{T}^i$ is generated, we will then call the pushing strategy to compute the first action $\bm{u}_0$ and investigate its feasibility lazily.
If $\bm{u}_0$ is not feasible (i.e., the starting position $P_0 \in \mathcal{W}$ for the pusher to perform $\bm{u}_0$ is occluded by other objects),
$\mathcal{T}^i$ cannot be achieved by the robot execution and will be set as $null$.
Otherwise, $\mathcal{T}^i$ will be simulated by the object-centric physics $\Pi$ to predict the outcome arrangement of objects $\bm{s}_o$, which will be added as a new node to expand the tree.

Note that the feasibility check for pushing actions considers only whether the starting position of a push is occluded by other objects or not.
We check only the first $\bm{u}_0$ as it requires the robot to reach the position $P_0$ from above the workspace (as will be described in Sec.~\ref{sec:action}), where undesired collisions with the top of objects may occur.
Subsequent pushing actions are assumed feasible, as they involve only in-plane motions of the robot.
Other feasibilities with respect to robot kinematics (e.g., joint limits) are not verified at this stage, 
since the robot configuration is not yet determined during object-centric planning. 
While this may lead to discrepancies during actual execution (e.g., the robot may be unable to execute a later action due to kinematic constraints and must skip it or replan, as detailed in Sec.~\ref{sec:action}),
in practice we find that the planner remains effective and reliably completes long-horizon tasks.

\begin{figure}[t]
\begin{algorithm}[H]
\small
\caption{ExpandTree($\cdot$)}
    \begin{algorithmic}[1]
        \Require A sampled tree node $n$, the index of the activated object $i$
        \Ensure The outcome arrangement $\bm{s}_o \in \mathcal{Q}^O$, the trajectory $\mathcal{T}^{i}$ of the activated object generated for exploration
        \State $\bm{s} \gets n.\bm{s}$ \hfill \Comment{Arrangement of the Node $n$}
        \State $\bm{s}^i = \left(x^i, y^i, \theta^i\right) \gets \Call{GetObjectState}{\bm{s}, i}$
        \If{$\Call{Uniform}{0, 1} > p_{A^*}$ or $i \in n.\mathcal{B}$} \hfill \Comment{\emph{Mode I}}
            \State $\mathcal{T}^i \gets \text{A Straight Line with Angle $\gamma$}$
        \Else \hfill \Comment{\emph{Mode II}}
            \State $\mathcal{T}^i \gets \Call{Soft-A*}{\bm{s}, i}$ \hfill \Comment{Sec.~\ref{sec:a_star}}
            \State $n.\mathcal{B} \gets n.\mathcal{B} \cup \{i\}$
        \EndIf
        \State $\hat{\bm{s}}_1^i \gets \text{First Waypoint of } \mathcal{T}^i$
        \State $\bm{u}_0 \gets \Call{PushStrategy}{\bm{s}^i, \hat{\bm{s}}_1^i}$ \hfill \Comment{Sec.~\ref{sec:uno_push}}
        \If{not $\Call{Feasible}{\bm{u}_0}$}
            \State \Return $null$
        \EndIf
        \State $\bm{s}_o \gets \Pi\left(\bm{s}, i, \mathcal{T}^i \right)$ \hfill \Comment{Simulate via $\Pi$}
        \State \Return $\bm{s}_o$, $\mathcal{T}^i$
    \end{algorithmic}
    \label{alg:expand_tree}
\end{algorithm}
\end{figure}

\subsection{Goal-Oriented Trajectory Planned by Soft-$A^*$}
\label{sec:a_star}

As one of the core designs of our OCP planner, \emph{Mode II} explores the object arrangement in a more goal-oriented manner compared to the random and local $\emph{Mode I}$.
The trajectory $\mathcal{T}^i$ generated by \emph{Mode II} intends to move the activated object $i$ to its optimal position that minimizes the heuristic cost $h(\bm{s})$,
while trying to avoid collisions with other objects to prevent loss of rearrangement progress.
An $A^*$ algorithm can efficiently plan such a collision-free trajectory with the shortest path length.
However, a collision-free trajectory may rarely exist, especially in a highly packed setup.
To this end, we relax the collision-free requirements of the original $A^*$ algorithm to propose the Soft-$A^*$, which allows the activated object $i$ to have insignificant collisions with other objects.
The Soft-$A^*$ plans the trajectory $\mathcal{T}^i$ by minimizing a cost that takes both the path length and potential collisions into account.
While our Soft-$A^*$ is fundamentally based on an $A^*$ algorithm with a modified cost function, 
it differs from traditional $A^*$ and its variants, which seek collision-free paths among static obstacles. 
In contrast, our Soft-$A^*$ is motivated by the nature of nonprehensile multi-object rearrangement tasks, where object-object interactions are often essential and unavoidable.

\begin{figure}[t]
    \centering
    \begin{tikzpicture}
        \node[anchor=south west,inner sep=0] at (0,0){\includegraphics[width=0.9\columnwidth]{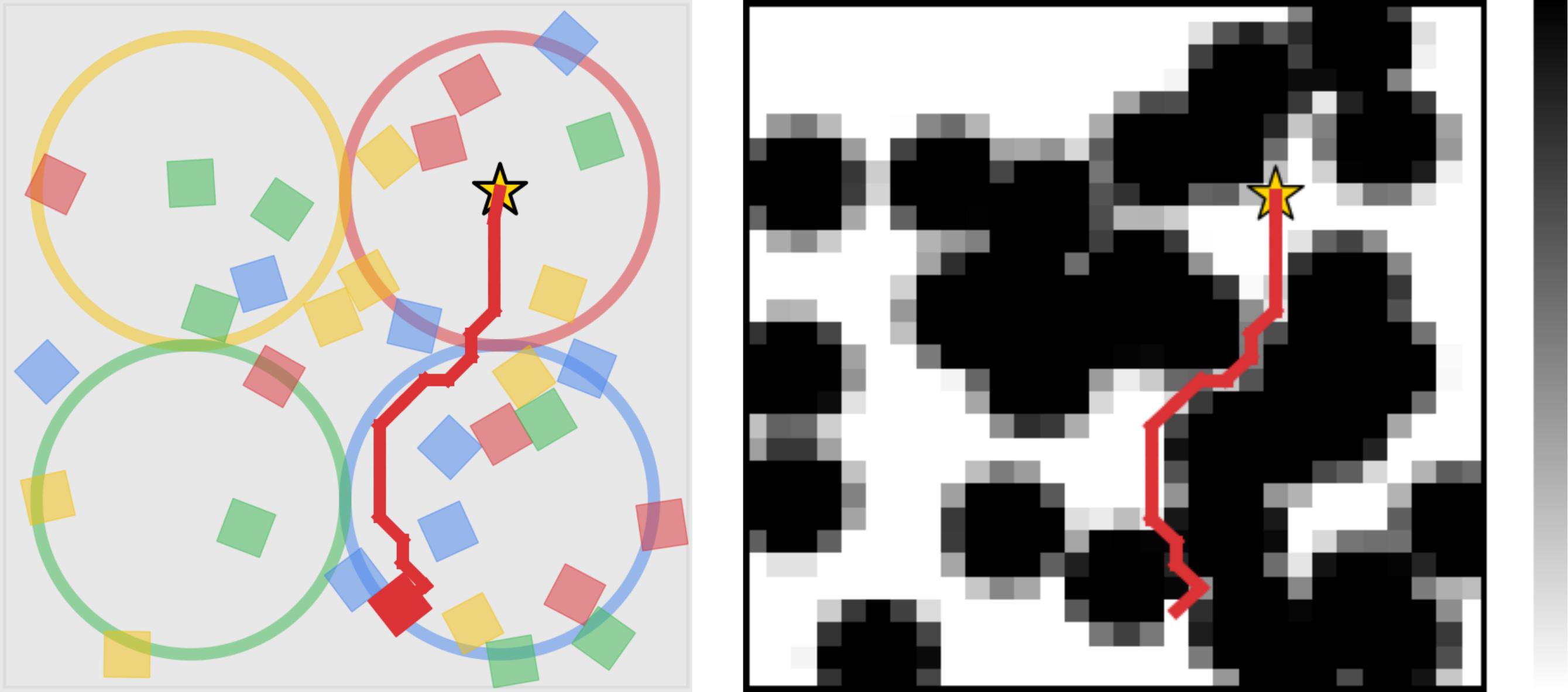}};
        \node[anchor=west, align=left] at (8, 3.5) {1.0};
        \node[anchor=west, align=left] at (8, 0) {0};
        \node[anchor=west, align=left] at (8, 1.75) {0.5};
    \end{tikzpicture}
    \caption{The grid representation of the workspace (right) for the activated red object (highlighted) in the scene (left).
    The values of grid cells, ranging from $0$ to $1$, evaluate the collisions between the activated object and other objects.
    The path (red lines) planned by the Soft-$A^*$ algorithm on the grid map will move the activated object from its current position to the optimal position (the star mark in the left figure) at the center of its goal region (the red circle).}
    \label{fig:a_star}
\end{figure}

As shown in Fig.~\ref{fig:a_star}, we discretize the workspace $\mathcal{W}$ into a finite grid map $\mathcal{I}^i \in \mathbb{R}^{W \times H}$ with a width $W$ and height $H$.
The size of a grid cell, denoted by $\Delta$, is a hyperparameter used to specify the map's resolution.
Each cell of the grid, denoted by $g \in \mathcal{I}^i$, has a coordinate $g.\bm{p} = (g.x, g.y)$ defined by the position of the cell's center point.
Each cell also has a value, denoted by $g.v \in [0, 1]$, which measures the potential collisions between the activated object $i$ and other objects.
We implement the evaluation of cell value by a linear function in terms of the distance between objects, as expressed in Eq.~\eqref{eq:g_v}.
\begin{subequations}
\begin{align}
    g.v &= 
    \begin{cases}
        1 & d_g < C_{\min} \\
        0 & d_g > C_{\max} \\
        (C_{\max} - d_g) / (C_{\max} - C_{\min}) & \text{otherwise}
    \end{cases} \label{eq:g_v} \\
    d_g &= \min_{j\in\{1, \cdots, N\}, j\neq i} \lVert g.\bm{p} - \bm{p}^j \rVert \label{eq:d_g}
\end{align}
\end{subequations}
where $d_g$ in Eq.~\eqref{eq:d_g} is the distance between a cell $g$ and the nearest object (excluding the activated object $i$) to this cell.
$C_{\min}$ and $C_{\max}$ are two parameters specified based on the size of the objects.
When $d_g < C_{\min}$, we consider the collision for the activated object $i$ (when at the position of the cell $g$) to be unacceptable, and thus set the cell value to $1$; 
when $d_g > C_{\max}$, the activated object is guaranteed not to collide with other objects, and the cell value is set to $0$; 
when $d_g$ decreases from $C_{\max}$ to $C_{\min}$, the cell value will increase linearly from $0$ to $1$ indicating an increase in potential collisions between the activated object and other objects.

Then we run an $A^*$ algorithm on the grid map $\mathcal{I}^i$ to find a path for the activated object, represented by the sequence of cells $\{g_1, g_2, \cdots, g_*\}$ it traverses.
The path starts from the current position and ends up at the optimal cell $g_*$ of the activated object.
The optimal cell $g_*$ is the cell that minimizes the heuristic cost $h(\cdot)$ when placing the activated object at this cell.
For many simple-formed heuristic functions, analytically finding the optimal $g_*$ is easy;
otherwise, we can brute-force search over the entire grid map to find $g_*$.
In practice, we find a resolution $\Delta$ similar to object size works well enough.
Thus, the size of the grid is usually small, making the brute-force search computationally cheap.

An $A^*$ algorithm conventionally needs two functions:
1) a cost function $c_{A*}(\cdot, \cdot)$ for two adjacent cells,
and 2) a heuristic function $h_{A^*}(\cdot)$ that underestimates the cost from a cell to the goal $g_*$.
Our Soft-$A^*$ aims to trade-off between the path length and potential collisions along the path. 
As such, we propose to use a cost function composed of both terms. 
Given a grid cell $g$ and one of its adjacent cell $g'$, the cost is evaluated by Eq.~\eqref{eq:c_astar}:
\begin{equation}
\begin{aligned}
    c_{A^*}(g, g') &= d(g, g') + \Delta \cdot g'.v\\
    d(g, g') &= \begin{cases}
        \lVert g.\bm{p} - g'.\bm{p} \rVert & g'.v < 1 \\
        \infty & \text{otherwise}
    \end{cases}
\end{aligned}
\label{eq:c_astar}
\end{equation}
The first term $d(g, g')$ is the distance between the two cells $g$ and $g'$; the second term $g'.v$ is a cell value that measures collision, which is scaled by the cell size $\Delta$ to have a comparable magnitude with the first term.
Importantly, when $g'$ has an unacceptable collision (i.e., $g'.v = 1$), the distance $d(g, g')$ is set to infinity to prevent the path from passing through $g'$.
The accumulated $c_{A*}(\cdot, \cdot)$ along the entire path, called the total cost, will be minimized by the Soft-$A^*$.

The heuristic function of our Soft-$A^*$ is simply the straight-line distance from a cell $g$ to the goal cell $g_*$, i.e., $ h_{A^*}(g) = \lVert g.\bm{p} - g_*.\bm{p} \rVert$.
Note that $h_{A^*}(g)$ strictly underestimates the cost from $g$ to $g_*$, which ensures the optimality of the found path.
If the total cost of the found path $\{g_1, g_2, \cdots, g_*\}$ is finite, we will generate a trajectory for the activated object by extracting the coordinates of each cell throughout the path and assuming the orientation of the object is not changed, 
i.e., $\mathcal{T}^i = \{\left(g_1.\bm{p}, \theta^i\right), \left(g_2.\bm{p}, \theta^i\right), \cdots, \left(g_*.\bm{p}, \theta^i \right)\}$.
This $\mathcal{T}^i$ will be returned to expand the motion tree under \emph{Mode II}.
Otherwise, when the total cost of the found path is infinite, we will set $\mathcal{T}^i = null$ to abort it.

\section{Robot Execution}
\label{sec:action}

After the OCP has planned object trajectories $\{\left(i_d, \mathcal{T}^{i_d} \right)\}_{d=1}^D$ in Sec.~\ref{sec:method}, each trajectory $\mathcal{T}^{i_d}$ needs to be realized by the real robot execution.
For each planned trajectory $\mathcal{T}^{i_d} = \{\hat{\bm{s}}_1^{i_d}, \cdots, \hat{\bm{s}}_k^{i_d}, \cdots, \hat{\bm{s}}_K^{i_d}\}$ of an activated object $i_d$, 
the robot will specify the activated object as the target object for pushing, and push it to reach each waypoint $\hat{\bm{s}}^{i_d}_k \in SE(2)$ in the trajectory one by one via a closed-loop pushing strategy (e.g., the UNO Push framework~\cite{wang2024uno}), as outlined in Alg.~\ref{alg:execute}.
Specifically, the robot needs to inspect the first pushing action $\bm{u}_0$ generated by the pushing strategy.
If the starting position $P_0 \in \mathcal{W}$ associated with $\bm{u}_0$ is occluded by objects, the robot will not be able to insert its pusher to this position for subsequent manipulation of the object $i_d$.
In this case, the robot will stop the current execution and skip to manipulate the next object.
Otherwise, the robot will move its pusher to $P_0$ from above the workspace;
and then by reaching each waypoint $\hat{\bm{s}}^i_k$, $k = 1, \cdots, K$ of the planned trajectory $\mathcal{T}^{i_d}$, the robot can manipulate the object $i_d$ through planar pushing (i.e., the pusher always moves in parallel to the workspace plane) to follow $\mathcal{T}^{i_d}$.
For implementing the robot execution, we control the motion of the pusher by commanding Cartesian velocities $\bm{v} \in se(3)$.
The Cartesian $\bm{v}$ is then projected into the robot's configuration space to generate a robot control $\bm{a} \in \mathcal{A}$ for commanding the robot.
In this work, we use the robot's joint velocity as control, i.e., $\bm{a} = \dot{\bm{q}} \in \mathbb{R}^M$, which is generated via null-space projection of the robot Jacobian, as detailed in Eq.~\eqref{eq:jacob_proj}:
\begin{equation}
\label{eq:jacob_proj}
    \bm{a} = \dot{\bm{q}} = J^\dagger \cdot \bm{v} + \lambda \cdot \left(\mathbb{I} - J^\dagger J \right) \dot{\bm{q}}_{null}
\end{equation}
where $\dot{\bm{q}}_{null} \in \mathbb{R}^M$ is the joint velocity for null-space motion, to improve the motion quality of the robot (e.g., moving away from singularity and joint limits);
we compute $\dot{\bm{q}}_{null}$ by taking the gradient of certain quality measures (e.g., the manipulability and a distance-based cost related to joint limits)~\cite{haviland2023manipulator};
$\dot{\bm{q}}_{null}$ is projected via $\left(\mathbb{I} - J^\dagger J \right)$ to the Jacobian's null space for not affecting the desired Cartesian behavior of the robot.

\begin{figure}[t]
\begin{algorithm}[H]
\small
\caption{Execute($\cdot$)}
    \begin{algorithmic}[1]
        \Require The current arrangement $\bm{s} \in \mathcal{Q}^O$,
        the index $i$ and the planned trajectory $\mathcal{T}^{i} = \{\hat{\bm{s}}_k^{i}\}_{k=1}^K$ of the activated object
        \State $\bm{u}_0 = (\alpha_0, \beta_0) \gets \Call{PushStrategy}{\bm{s}^{i}, \hat{\bm{s}}_1^i}$ \hfill \Comment{Sec.~\ref{sec:uno_push}}
        \State $P_0 \gets \Call{StartingPosition}{\bm{s}^i, \bm{u}_0}$
        \If{$\Call{Occluded}{P_0}$}
            \State \Return
        \EndIf
        \State $\Call{MovePusherTo}{P_0}$ \hfill \Comment{Above the Workspace}
        \For{$k = 1, \cdots, K$}
            \While{$\bm{s}^i$ not reaching $\hat{\bm{s}}^i_k$}
                \State $\bm{u} \gets \Call{PushStrategy}{\bm{s}^i, \hat{\bm{s}}^i_k}$ \hfill \Comment{Sec.~\ref{sec:uno_push}}
                \State $\Call{Push}{\bm{u}, d_{push}}$ \hfill \Comment{Planar Push by a Distance $d_{push}$}
                \State $\bm{s} \gets \Call{ObserveObjects}{\null}$ \hfill \Comment{via Real-time Sensing}
            \EndWhile
        \EndFor
    \end{algorithmic}
    \label{alg:execute}
\end{algorithm}
\end{figure}

Throughout the execution, the robot constraints must be satisfied to ensure that future execution is still feasible.
With the same frequency as the low-level controller, the robot will monitor the following three events by real-time sensor readings:
1) self-collision; 2) joint limit violation; and 3) singularity measured by the volume-based manipulability $\sqrt{\mathrm{det}\left(J J^{\top} \right)}$.
If any event is about to occur, we will stop the robot execution, safely move the robot back to the previous configuration, and skip to manipulate the next target object.
This strategy ensures the robot complies with its constraints in a lazy manner.
Furthermore, to make sure no objects will be pushed outside the workspace $\mathcal{W}$ due to the inaccuracy of the modeled physics and real-world uncertainties,
we also monitor the distances between each object and the workspace boundary.
When an object gets too close to the workspace boundary, we will use the same pushing strategy to push that object back toward the center of the workspace.

\section{Experiments}
\label{sec:experiemnts}

\begin{figure*}[t]
    \centering
    \begin{tikzpicture}
        \node[anchor=south west,inner sep=0] at (0,0){\includegraphics[width=\linewidth]{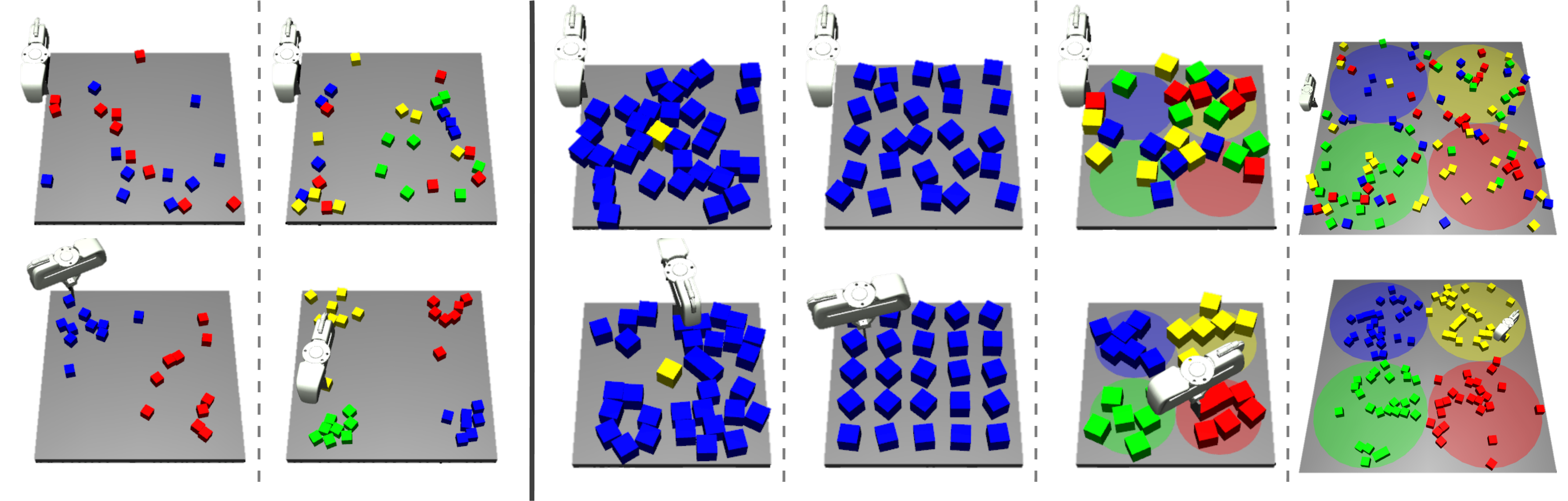}};
        \node[anchor=center, align=center] at (3, 0) {Sorting without Goal Regions};
        \node[anchor=center, align=center] at (7.8, 0) {Singulate};
        \node[anchor=center, align=center] at (10.7, 0) {Separate};
        \node[anchor=center, align=center] at (15, 0) {Sorting with Goal Regions};

        \node[anchor=center, align=center] at (1.65, 5.7) {$L = 2$, $N = 20$};
        \node[anchor=center, align=center] at (4.55, 5.7) {$L = 4$, $N = 30$};
        \node[anchor=center, align=center] at (7.8, 5.7) {$N = 33$};
        \node[anchor=center, align=center] at (10.7, 5.7) {$N = 25$};
        \node[anchor=center, align=center] at (13.6, 5.7) {$L = 4$, $N = 24$};
        \node[anchor=center, align=center] at (16.5, 5.7) {$L = 4$, $N = 100$};
    \end{tikzpicture}
    \caption{The evaluated rearrangement tasks solved by our OCP planner in MuJoCo simulation with a floating gripper pusher.
    The task scenes have different numbers of objects ($N$) and object classes ($L$).
    The top figures show the initial configuration of randomly placed objects, and the bottom figures show the final configurations when the tasks are completed. }
    \label{fig:tasks}
\end{figure*}

Extensive experiments were conducted in both simulation and on a physical robot manipulator to evaluate the performance of our proposed object-centric planner (OCP) against selected baselines.
Our evaluation focuses on two key aspects:
1) Planning efficiency, 
measured by the average planning time for task completion (lower is better);
and 2) Action effectiveness, 
assessed by the average number of actions and average execution time (for real-world experiments) to complete a task.
Our planner was implemented with the Box2D physics engine~\footnote{Box2D, A 2D Physics Engine for Games: \url{https://box2d.org/}} to approximate the object-centric transition function $\Pi$ defined in Sec.~\ref{sec:prob}.
Unless otherwise specified,
all evaluations were run in Python with a single thread on a 3.4 GHz AMD Ryzen 9 5950X CPU.
For simulation evaluation, we used a floating gripper in the MuJoCo~\cite{todorov2012mujoco} environment to execute actions generated by the planner, as shown in Fig.~\ref{fig:tasks}.
The floating gripper is allowed to freely navigate parallel to the workspace, to interact with objects using an attached fence pusher.
The gripper can also teleport to switch to manipulate a different object.
For real-world experiments, we evaluated our planner on a 7-DoF Franka Emika Panda robot manipulator.

\subsection{Planar Pushing-based Rearrangement Tasks}
\label{sec:tasks}
We selected challenging rearrangement tasks from two state-of-the-art works~\cite{huang2019large, song2020multi} for evaluation, 
by considering the variety and large scale of entailed tasks.
For example, some tasks (e.g., singulation) require only local rearrangement of a target object, whereas others (e.g., sorting) involve global rearrangement and relative reconfiguration of all objects.
All evaluated tasks are visualized in Fig.~\ref{fig:tasks}.
We generally categorize all tasks into two types, based on how the task goals are defined differently:

\subsubsection{Tasks without Explicit Goal Definition}
Such tasks do not require an explicitly defined goal pose or region for each object.
Instead,
as long as the relative reconfiguration of object states meets certain requirements, the task is considered successful.
One example of such tasks is \emph{Object Sorting without Goal Regions} investigated in~\cite{song2020multi}.
As shown in Fig.~\ref{fig:tasks} (left), the robot is tasked to separate objects of $L$ different classes into clusters.
No specific goal pose or region is assigned to any object.
By observing the object arrangement $\bm{s} \in \mathcal{Q}^O$, 
a convex hull $\mathbf{CH}_j(\bm{s}) \subset \mathbb{R}^2$ is computed for each class $j \in \{1, \cdots, L\}$ to enclose objects of that class (i.e., covering geometric shapes of all objects belonging to class $j$).
The task is considered successful if the minimum distance between any pair of convex hulls is larger than a threshold $\epsilon_d$.
Formally, the goal criterion is satisfied (i.e., $g(\bm{s}) = 1$) when $\forall i, j \in \{1, \cdots, L\}$:
\begin{equation}
    \min_{i \neq j} \mathrm{dist}\left(\mathbf{CH}_i(\bm{s}), \mathbf{CH}_j(\bm{s}) \right) > \epsilon_d
\end{equation}
To solve this task, we equip our object-centric planner with a heuristic function $h(\cdot)$ similar to the reward function in~\cite{song2020multi}.
A lower heuristic cost will be expected when intra-class objects cluster more tightly and inter-class objects are more separated.

\subsubsection{Tasks with Explicit Goal Definition}
This type of task requires an explicit goal region $\mathcal{G}_i \subset \mathcal{W}$ for each movable object $i \in \{1, \cdots, N\}$,
with its centroid position denoted by $\bm{p}^{\mathcal{G}_i} = \left(x^{\mathcal{G}_i}, y^{\mathcal{G}_i} \right)$.
In different tasks, the goal regions of some objects can be either distinct or overlapping.
The task is considered successful when each object is relocated within its designated goal region, formally defined as:
\begin{equation}
    \forall i \in \{1, \cdots, N\} : \bm{p}^i = \left(x^i, y^i \right) \in \mathcal{G}_i
\end{equation}
where $\bm{p}^i = \left(x^i, y^i \right)$ is the position of the $i$-th object.
The heuristic function used by our object-centric planner is simply the squared distances between objects and their corresponding goal regions:
\begin{equation}
    h(\bm{s}) = \sum_{i \in \{1, \cdots, N\}} \mathbbm{1} \left\{\bm{p}^i \notin \mathcal{G}_i \right\} \frac{\lVert \bm{p}^i - \bm{p}^{\mathcal{G}_i} \rVert ^ 2}{r_{\mathcal{G}_i}^2}
\end{equation}
where $r_{\mathcal{G}_i}$ denotes the size of the goal region $\mathcal{G}^i$ (e.g., a radius if $\mathcal{G}_i$ is a circle); 
$\mathbbm{1}\left\{\bm{p}^i \notin \mathcal{G}_i \right\}$ is an indicator function that excludes objects already within their goal regions.
When using heuristic gradients to guide exploration (in Sec.~\ref{sec:obj_sampling}), this indicator function helps our planner focus on exploring unsolved objects (i.e., objects not at their goals) without wasting time on already solved objects.

The tasks from~\cite{huang2019large} can all be categorized into this type:
\begin{enumerate}[label=\alph*)]
    \item \emph{Singulate:} 
    This task requires a target object $o \in \{1, \cdots, N\}$ to be isolated from the other objects $i \neq o$.
    Specifically, the target object needs to be relocated inside a small circular goal region $\mathcal{G}^o$ at the center of the workspace, while other objects are moved near one of the four corners.
    
    \item \emph{Separate:}
    The workspace is divided into multiple grid cells, each containing a tiny circular goal region.
    Each object must be placed inside a distinct goal region.
    Since objects are not unique, the assignment of objects to their goal regions is dynamically determined.

    \item \emph{Sorting with Goal Regions:}
    There are four classes of objects, visually represented by different colors.
    The objects of the same class share the same goal region, which is a circle located near one corner of the workspace.
    The task requires the objects of different classes to be moved to their corresponding goal regions.
\end{enumerate}

In certain tasks (e.g., Separate), some objects are not unique, and their corresponding goal regions are interchangeable.
In such cases, similar to~\cite{huang2019large}, the assignment of objects to goal regions is dynamically determined.
Specifically, 
let $\mathcal{D} \subset \{1, \cdots, N\}$ denote the set of non-unique objects, and let $\mathcal{SG} = \{\mathcal{G}^i : i \in \mathcal{D} \}$ denote the set of distinct goal regions for these objects.
Note that $\lvert \mathcal{D} \rvert = \lvert \mathcal{SG} \rvert$.
An assignment function $A: \mathcal{D} \mapsto \mathcal{SG}$ bijectively assigns each object $i \in \mathcal{D}$ to its corresponding goal region $A(i) \in \mathcal{SG}$.
This assignment is dynamically determined by minimizing the summed distances between objects and their assigned goal region, i.e., $\sum_{i \in \mathcal{D}} \mathrm{dist}\left(\bm{p}^i, A(i)\right)$.

\subsection{Selected Baselines for Evaluation}
\label{sec:baselines}
As extensively discussed in Sec.~\ref{sec:intro} and Sec.~\ref{sec:relatedwork},
many relevant approaches are proposed to address planar nonprehensile rearrangement, 
including sampling-based~\cite{king2015nonprehensile, haustein2015kinodynamic} and end-to-end learning-based methods~\cite{eitel2020learning}, 
as well as robot-centric~\cite{ren2022rearrangement} and object-centric approaches~\cite{barry2013manipulation, saxena2023planning}.
However, many of these methods are designed for specific scenarios (e.g., confined spaces) or smaller-scale problems (e.g., pre-grasp manipulation for object retrieval~\cite{dogar2012physics}),
making them less adaptable to larger-scale tasks with diverse objectives as described in Sec.~\ref{sec:tasks}.
Among them, we select three relevant baselines that align with our task setups and can be reasonably re-implemented for evaluation.
For fair comparison,
we re-implemented each baseline and adapted its implementation to meet our task requirements, as detailed below:
\begin{itemize}[wide]
    \item Hybrid-RRT (H-RRT)~\cite{king2016rearrangement}: A sampling-based approach based on a kinodynamic RRT, where actions are hybridly sampled between robot-centric actions and object-centric pushing primitives.
    For convenience, we refer to this method as Hybrid-RRT (abbreviated as H-RRT) throughout the paper.
    To improve sampling efficiency for large-scale tasks, 
    we introduce a goal bias to RRT with a probability of $0.2$ for sampling a goal state for tree expansion,
    if goal configurations are explicitly given by the task.
    In addition, we made the following key adaptations:
    1) The original planner is fully open-loop and prone to failure in long-horizon tasks due to modeling inaccuracies. 
    To address this, we adapted the implementation to perform limited-horizon planning, execute the best candidate node, sense the real-world state, and then replan iteratively.
    2) The original planner assumes a single target object.
    To extend it to handle multiple target objects,
    when sampling an object-centric action,
    we randomly sample one object that is not at its goal and treat it as the current target. 
    If no goal is specified, we sample a random object to push.
    3) For fair comparison, we use the same Box2D physics engine for dynamics propagation.
    
    \item ILS~\cite{huang2019large}: An Iterative Local Search (ILS) algorithm that locally explores and optimizes robot actions that reduce the distance of objects to their goal regions.
    The original implementation also uses Box2D.
    We made the following key adaptations: 
    1) Since the algorithm is open-loop, 
    we modify it to execute each locally searched trajectory, followed by sensing and replanning to close the loop.
    2) The original ILS assumes goal regions are given and uses them to guide object and (greedy) action sampling.
    For tasks without explicit goal regions,
    we instead randomly sample an object, and action sampling becomes fully random.
    3) For tasks without explicit goal regions,
    the original distance-based cost is no longer applicable;
    in such cases, we replace it with the same heuristic cost used in our planner.

    \item kdRRF~\cite{ren2023kinodynamic}: A forest-based kinodynamic planner that enables concurrent exploration from different subspaces of the problem to find more task-efficient motions.
    For a fair comparison, we replaced the MuJoCo-based physics model used in the original implementation of kdRRF with the Box2D physics engine.
\end{itemize}

In addition to the re-implemented baselines, 
we also include a comparison with the method from~\cite{song2020multi} when evaluating on the same tasks used in their study.
Their approaches uses a Monte Carlo Tree Search (MCTS)-based planner guided by a learned rollout policy (i.e., a deep neural network trained offline using image data).
We found that this method requires substantial dataset collection and retraining to adapt to new rearrangement tasks or setups.
For this reason, we did not re-implement it for new tasks and instead report their original results as a reference.
As will be shown by the experiments in Sec.~\ref{sec:sim_exp} and~\ref{sec:real_world_exp}, even compared to their original parallelized implementation and fine-tuned policy with data, our planner achieves comparable success on similar tasks, without requiring any data collection or retraining.

\subsection{Analysis of Parameter Selection and Ablation Study}

We first conducted simulation experiments to analyze how the performance of our OCP planner is affected by different choices of key parameters: 
a) the maximum tree size $S_{\max}$, 
and b) the probability $p_{A^*} $ of selecting \emph{Mode II} (i.e., Soft-$A^*$) for exploration.
Other parameters, such as the pushing distance $d_{push}$, the length range of a sampled \emph{Mode I} motion $[l_{\min}, l_{\max})$, and the grid resolution $\Delta$ of Soft-$A^*$ can be selected more intuitively relative to the object and workspace sizes and do not require extensive fine-tuning for functionality.
We chose the representative task, \emph{Sorting with Goal Regions}, to evaluate all parameter settings.
In this task, $32$ objects belonging to $4$ classes ($8$ per class) must be sorted, following a similar setup to Fig.~\ref{fig:tasks}.
The workspace was a $0.6m \times 0.6m$ square, with circular goal regions (for the $4$ classes) centered at $(\pm 0.135, \pm 0.135)m$,
each with a radius of $0.135m$.
For each parameter setting, we ran $100$ trials with the same initial object configuration to collect performance statistics.

\begin{figure}[t]
    \centering

    \begin{minipage}[b]{\linewidth}
        \centering
\begin{tikzpicture}

\definecolor{mycolor1}{rgb}{0.23529,0.72941,0.32941}%
\definecolor{mycolor2}{rgb}{0.95686,0.76078,0.05098}%
\definecolor{mycolor3}{rgb}{0.85882,0.19608,0.21176}%
\definecolor{mycolor4}{rgb}{0.28235,0.52157,0.92941}%
\definecolor{mycolor5}{rgb}{1.00000,0.54902,0.00000}%

\begin{axis}[
width=0.85\columnwidth,
height=0.3\columnwidth,
scale only axis,
grid=both,
grid style={line width=.1pt, draw=gray!20},
major grid style={line width=.2pt,draw=gray!50},
xmin=0, xmax=1,
xtick={0, 0.2, 0.5, 0.8, 1.0},
ymin=100, ymax=250,
ytick={100, 150, 200, 250},
minor y tick num=4,
ylabel={number of actions},
title style={font=\bfseries},
axis x line*=bottom,
axis y line*=left,
legend style={legend cell align=left, align=left, legend columns = 3, fill=none, draw=none, at={(0.98,0.85)},anchor=east}
]
\path [fill=mycolor1, fill opacity=0.3]
(axis cs:0,233.624409486255)
--(axis cs:0,210.075590513745)
--(axis cs:0.2,178.41863723038)
--(axis cs:0.5,173.540433434123)
--(axis cs:0.8,168.151597427251)
--(axis cs:1,165.271603568969)
--(axis cs:1,179.788396431031)
--(axis cs:1,179.788396431031)
--(axis cs:0.8,187.368402572748)
--(axis cs:0.5,191.939566565877)
--(axis cs:0.2,195.64136276962)
--(axis cs:0,233.624409486255)
--cycle;

\path [fill=mycolor2, fill opacity=0.3]
(axis cs:0,210.144811044945)
--(axis cs:0,187.355188955055)
--(axis cs:0.2,164.269414196702)
--(axis cs:0.5,151.23433203254)
--(axis cs:0.8,149.14625357143)
--(axis cs:1,153.270598062593)
--(axis cs:1,168.569401937407)
--(axis cs:1,168.569401937407)
--(axis cs:0.8,161.83374642857)
--(axis cs:0.5,164.02566796746)
--(axis cs:0.2,178.470585803298)
--(axis cs:0,210.144811044945)
--cycle;

\path [fill=mycolor3, fill opacity=0.3]
(axis cs:0,186.978929123718)
--(axis cs:0,168.041070876282)
--(axis cs:0.2,145.820238118157)
--(axis cs:0.5,144.029875080305)
--(axis cs:0.8,140.299870771932)
--(axis cs:1,135.924152728092)
--(axis cs:1,146.635847271908)
--(axis cs:1,146.635847271908)
--(axis cs:0.8,153.600129228068)
--(axis cs:0.5,159.790124919695)
--(axis cs:0.2,161.279761881843)
--(axis cs:0,186.978929123718)
--cycle;

\path [fill=mycolor4, fill opacity=0.3]
(axis cs:0,172.703908680938)
--(axis cs:0,154.556091319062)
--(axis cs:0.2,138.08914238999)
--(axis cs:0.5,132.805842258769)
--(axis cs:0.8,136.888254970763)
--(axis cs:1,130.370876158976)
--(axis cs:1,141.200552412452)
--(axis cs:1,141.200552412452)
--(axis cs:0.8,148.070928702706)
--(axis cs:0.5,143.854157741231)
--(axis cs:0.2,149.13085761001)
--(axis cs:0,172.703908680938)
--cycle;

\path [fill=mycolor5, fill opacity=0.3]
(axis cs:0,165.042983128598)
--(axis cs:0,150.877016871402)
--(axis cs:0.2,124.157023495714)
--(axis cs:0.5,118.258909704174)
--(axis cs:0.8,116.690449708379)
--(axis cs:1,112.238521524776)
--(axis cs:1,120.623016936762)
--(axis cs:1,120.623016936762)
--(axis cs:0.8,124.420661402732)
--(axis cs:0.5,127.284568556696)
--(axis cs:0.2,133.822774484084)
--(axis cs:0,165.042983128598)
--cycle;

\addplot [semithick, mycolor1, mark=*, mark size=3, mark options={solid}]
table {%
0 221.85
0.2 187.03
0.5 182.74
0.8 177.76
1 172.53
};
\addlegendentry{20}
\addplot [semithick, mycolor2, mark=*, mark size=3, mark options={solid}]
table {%
0 198.75
0.2 171.37
0.5 157.63
0.8 155.49
1 160.92
};
\addlegendentry{50}
\addplot [semithick, mycolor3, mark=*, mark size=3, mark options={solid}]
table {%
0 177.51
0.2 153.55
0.5 151.91
0.8 146.95
1 141.28
};
\addlegendentry{100}
\addplot [semithick, mycolor4, mark=*, mark size=3, mark options={solid}]
table {%
0 163.63
0.2 143.61
0.5 138.33
0.8 142.479591836735
1 135.785714285714
};
\addlegendentry{200}
\addplot [semithick, mycolor5, mark=*, mark size=3, mark options={solid}]
table {%
0 157.96
0.2 128.989898989899
0.5 122.771739130435
0.8 120.555555555556
1 116.430769230769
};
\addlegendentry{500}
\end{axis}

\end{tikzpicture}
        \vspace{-5pt}
    \end{minipage}
    \begin{minipage}[b]{\linewidth}
        \centering
\begin{tikzpicture}

\definecolor{mycolor1}{rgb}{0.23529,0.72941,0.32941}%
\definecolor{mycolor2}{rgb}{0.95686,0.76078,0.05098}%
\definecolor{mycolor3}{rgb}{0.85882,0.19608,0.21176}%
\definecolor{mycolor4}{rgb}{0.28235,0.52157,0.92941}%
\definecolor{mycolor5}{rgb}{1.00000,0.54902,0.00000}%

\begin{axis}[
width=0.85\columnwidth,
height=0.3\columnwidth,
scale only axis,
grid=both,
grid style={line width=.1pt, draw=gray!20},
major grid style={line width=.2pt,draw=gray!50},
xmin=0, xmax=1,
xtick={0, 0.2, 0.5, 0.8, 1.0},
xlabel={$p_{A^*}$: probability of sampling a \emph{Mode II} motion},
ylabel={time per action},
ymin=0, ymax=1.2,
ytick={0, 0.5, 1.0, 1.5},
minor y tick num=4,
ylabel={time per action (seconds)},
title style={font=\bfseries},
axis x line*=bottom,
axis y line*=left,
legend style={legend cell align=left, align=left, legend columns = 3, fill=none, draw=none, at={(0.02,0.85)},anchor=west}
]

\path [fill=mycolor1, fill opacity=0.3]
(axis cs:0,0.0340560894391202)
--(axis cs:0,0.0241165740038836)
--(axis cs:0.2,0.0350151975643733)
--(axis cs:0.5,0.0483010868349944)
--(axis cs:0.8,0.0640098173370865)
--(axis cs:1,0.0764997051145831)
--(axis cs:1,0.0930352528321344)
--(axis cs:1,0.0930352528321344)
--(axis cs:0.8,0.0785016548810273)
--(axis cs:0.5,0.0596642693852954)
--(axis cs:0.2,0.0406272220270558)
--(axis cs:0,0.0340560894391202)
--cycle;

\path [fill=mycolor2, fill opacity=0.3]
(axis cs:0,0.0615543322644135)
--(axis cs:0,0.0519457732123719)
--(axis cs:0.2,0.0721828632210739)
--(axis cs:0.5,0.100743005178364)
--(axis cs:0.8,0.131184946482762)
--(axis cs:1,0.144435398829513)
--(axis cs:1,0.175695859479118)
--(axis cs:1,0.175695859479118)
--(axis cs:0.8,0.152544880612238)
--(axis cs:0.5,0.118528119396423)
--(axis cs:0.2,0.0848476093657448)
--(axis cs:0,0.0615543322644135)
--cycle;

\path [fill=mycolor3, fill opacity=0.3]
(axis cs:0,0.108757552808485)
--(axis cs:0,0.0962394080279717)
--(axis cs:0.2,0.130157222001647)
--(axis cs:0.5,0.18264581371598)
--(axis cs:0.8,0.224918776209894)
--(axis cs:1,0.255891240140518)
--(axis cs:1,0.297552506866692)
--(axis cs:1,0.297552506866692)
--(axis cs:0.8,0.269729275388409)
--(axis cs:0.5,0.210429948993021)
--(axis cs:0.2,0.148686937464539)
--(axis cs:0,0.108757552808485)
--cycle;

\path [fill=mycolor4, fill opacity=0.3]
(axis cs:0,0.206812984554707)
--(axis cs:0,0.180492993146729)
--(axis cs:0.2,0.246600190500358)
--(axis cs:0.5,0.337457209762252)
--(axis cs:0.8,0.388975715212434)
--(axis cs:1,0.438246465938127)
--(axis cs:1,0.512704135901229)
--(axis cs:1,0.512704135901229)
--(axis cs:0.8,0.477965418061179)
--(axis cs:0.5,0.395220284463978)
--(axis cs:0.2,0.285966013140183)
--(axis cs:0,0.206812984554707)
--cycle;

\path [fill=mycolor5, fill opacity=0.3]
(axis cs:0,0.501654538521363)
--(axis cs:0,0.445173604158352)
--(axis cs:0.2,0.596417238865102)
--(axis cs:0.5,0.772178736216993)
--(axis cs:0.8,0.88412708892531)
--(axis cs:1,0.924874723008151)
--(axis cs:1,1.09387332108997)
--(axis cs:1,1.09387332108997)
--(axis cs:0.8,1.03918623276678)
--(axis cs:0.5,0.900925437585208)
--(axis cs:0.2,0.691136646004543)
--(axis cs:0,0.501654538521363)
--cycle;

\addplot [semithick, mycolor1, mark=*, mark size=3, mark options={solid}]
table {%
0 0.0290863317215019
0.2 0.0378212097957146
0.5 0.0539826781101449
0.8 0.0712557361090569
1 0.0847674789733587
};
\addlegendentry{20}
\addplot [semithick, mycolor2, mark=*, mark size=3, mark options={solid}]
table {%
0 0.0567500527383927
0.2 0.0785152362934094
0.5 0.109635562287393
0.8 0.1418649135475
1 0.160065629154315
};
\addlegendentry{50}
\addplot [semithick, mycolor3, mark=*, mark size=3, mark options={solid}]
table {%
0 0.102498480418228
0.2 0.139422079733093
0.5 0.196537881354501
0.8 0.247324025799151
1 0.276721873503605
};
\addlegendentry{100}
\addplot [semithick, mycolor4, mark=*, mark size=3, mark options={solid}]
table {%
0 0.193652988850718
0.2 0.266283101820271
0.5 0.366338747113115
0.8 0.433470566636807
1 0.475475300919678
};
\addlegendentry{200}
\addplot [semithick, mycolor5, mark=*, mark size=3, mark options={solid}]
table {%
0 0.473414071339857
0.2 0.643776942434823
0.5 0.8365520869011
0.8 0.961656660846047
1 1.00937402204906
};
\addlegendentry{500}
\end{axis}

\end{tikzpicture}   
    \end{minipage}
    \vspace{-25pt}
    \caption{The statistics of the number of actions (top plot) and the average planning time per action (bottom plot) in terms of different parameter choices of $S_{\max}$ (plots in different colors) and $p_{A^*}$ (x-axis).  
    The shaded areas are the quarter standard deviations for the top plot; and the standard deviations for the bottom plot.}
    \label{fig:exp_analysis}
\end{figure}

The results, shown in Fig.~\ref{fig:exp_analysis},
illustrate the effects of different parameter choices.
Specifically, as the tree size increases with $S_{\max} = 20$, $50$, $100$, $200$, and $500$ (plotted in different colors), 
the planning time per action increases,
while the number of actions required to complete the task decreases.
This trend is intuitive: 
A larger $S_{\max}$ means the planner explores more object motions (i.e., more tree nodes) before selecting an action,
increasing computational effort per action;
at the same time, when $S_{\max}$ increases, the planner explores the problem space more extensively with more tree branches, which increases the probability of finding more optimal actions and results in a smaller number of actions needed for task completion.
When varying the probability $p_{A^*} = 0$, $0.2$, $0.5$, $0.8$, and $1.0$, we observed that a higher $p_{A^*}$ led to a longer planning time required for each action.
This is because a higher $p_{A^*}$ triggers more attempts of a \emph{Mode II} exploration, which requires an extra computational time of $A^*$ search.
However, 
higher $p_{A^*}$ also results in fewer needed actions.
This is because \emph{Mode II} motions are generally more effective than \emph{Mode I} motions in reducing the heuristic cost and progressing towards the task goal.

Overall,
the results have shown that a larger $S_{\max}$ and a higher $p_{A^*}$ generate more effective actions, which may potentially reduce the overall execution time by lowering the number of actions needed for execution.
In contrast, a smaller $S_{\max}$ and a lower $p_{A^*}$ can facilitate more reactive planning with shorter planning time per action, but at the cost of the optimality of generated actions.

Furthermore, with a tree size limit $S_{\max} = 50$,
we conducted an ablation study on the same sorting task to evaluate the importance of both exploration modes in our planner. 
Specifically, we compared the performance of our full OCP planner against three variants:
1) \emph{Mode II} (Soft-$A^*$) only;
2) \emph{Mode I} only (by setting $p_{A^*} = 0$);
3) using standard $A^*$ (i.e., only allowing strictly collision-free paths) in \emph{Mode II}.
We observed that the first variant (i.e., \emph{Mode II} only) was highly ineffective in cluttered scenarios (i.e., zero success rate), 
as the algorithm often got stuck in states where no solutions to the deterministic Soft-$A^*$ are feasible for any object.
In such cases, it failed to make progress and lacked a mechanism to escape such dead ends,
highlighting the necessity of \emph{Mode I} exploration to recover.
For the other two variants, we reported the performance in Fig.~\ref{fig:exp_ablation}.
As can be seen from the results,
compared to using \emph{Mode I} alone, our planner incurred slightly higher planning time due to the computational cost of Soft-$A^*$ in \emph{Mode II}, but produced more effective behavior with fewer actions;
compared to using standard $A^*$, our planner generated more optimal actions with lower planning time,
highlighting the benefits of incorporating our Soft-$A^*$.
Overall, the results show the importance of combining both exploration modes for effective and robust planning.

\begin{figure}[t]
    \centering
    \begin{minipage}{0.45\columnwidth}
\begin{tikzpicture}

\definecolor{mycolor1}{rgb}{0.23529,0.72941,0.32941}%
\definecolor{mycolor2}{rgb}{0.95686,0.76078,0.05098}%
\definecolor{mycolor3}{rgb}{0.85882,0.19608,0.21176}%
\definecolor{mycolor4}{rgb}{0.28235,0.52157,0.92941}%
\definecolor{mycolor5}{rgb}{1.00000,0.54902,0.00000}%

\begin{axis}[
width=0.7\columnwidth,
height=0.6\columnwidth,
scale only axis,
grid=both,
grid style={line width=.1pt, draw=gray!20},
major grid style={line width=.2pt,draw=gray!50},
xmin=-1, xmax=3,
ymin=0, ymax=30,
ylabel={planning time},
xmajorgrids=false,
xminorgrids=false,
xmajorticks=false,
xminorticks=false,
minor y tick num=4,
title style={font=\bfseries},
axis x line*=bottom,
axis y line*=left,
legend style={legend cell align=left, align=left, legend columns = 1, fill=none, draw=none, at={(0.98,0.75)},anchor=east}
]
\draw[draw=none,fill=mycolor3] (axis cs:-0.4,0) rectangle (axis cs:0.4,6.50656812906265);
\addlegendimage{only marks, mark=square*, fill=mycolor3}
\addlegendentry{\emph{Mode I} only}

\draw[draw=none,fill=mycolor2] (axis cs:0.6,0) rectangle (axis cs:1.4,10.7158100962639);
\addlegendimage{only marks, mark=square*, fill=mycolor2}
\addlegendentry{standard $A^*$}

\draw[draw=none,fill=mycolor4] (axis cs:1.6,0) rectangle (axis cs:2.4,9.878642578125);
\addlegendimage{only marks, mark=square*, fill=mycolor4}
\addlegendentry{OCP (Ours)}

\path [draw=black, very thick]
(axis cs:0,4.56238237788686)
--(axis cs:0,8.45075388023844);

\path [draw=black, very thick]
(axis cs:1,8.15694599529861)
--(axis cs:1,13.2746741972292);

\path [draw=black, very thick]
(axis cs:2,7.57079198832673)
--(axis cs:2,12.1864931679233);

\end{axis}

\end{tikzpicture}  
    \end{minipage}
    \begin{minipage}{0.45\columnwidth}
        \vspace{2pt}
\begin{tikzpicture}

\definecolor{mycolor1}{rgb}{0.23529,0.72941,0.32941}%
\definecolor{mycolor2}{rgb}{0.95686,0.76078,0.05098}%
\definecolor{mycolor3}{rgb}{0.85882,0.19608,0.21176}%
\definecolor{mycolor4}{rgb}{0.28235,0.52157,0.92941}%
\definecolor{mycolor5}{rgb}{1.00000,0.54902,0.00000}%

\begin{axis}[
width=0.7\columnwidth,
height=0.6\columnwidth,
scale only axis,
grid=both,
grid style={line width=.1pt, draw=gray!20},
major grid style={line width=.2pt,draw=gray!50},
xmin=-1, xmax=3,
ymin=0, ymax=300,
ylabel={number of actions},
xmajorgrids=false,
xminorgrids=false,
xmajorticks=false,
xminorticks=false,
minor y tick num=4,
title style={font=\bfseries},
axis x line*=bottom,
axis y line*=left,
legend style={legend cell align=left, align=left, legend columns = 3, fill=none, draw=none, at={(0.98,0.85)},anchor=east}
]
\draw[draw=none,fill=mycolor3] (axis cs:-0.4,0) rectangle (axis cs:0.4,221.85);

\draw[draw=none,fill=mycolor2] (axis cs:0.6,0) rectangle (axis cs:1.4,204.62);

\draw[draw=none,fill=mycolor4] (axis cs:1.6,0) rectangle (axis cs:2.4,182.74);

\path [draw=black, very thick]
(axis cs:0,174.752362054982)
--(axis cs:0,268.947637945018);

\path [draw=black, very thick]
(axis cs:1,160.37075593866)
--(axis cs:1,248.86924406134);

\path [draw=black, very thick]
(axis cs:2,145.941733736492)
--(axis cs:2,219.538266263508);

\end{axis}

\end{tikzpicture}  
    \end{minipage}
    \caption{
        The statistics of planning time (left) and number of actions (right) for different variants of our planner in the ablation study:
        only with \emph{Mode I} motions by setting $p_{A^*} = 0$ (red);
        our planner with Soft-$A^*$ replaced by a standard $A^*$ (yellow);
        our planner (blue).
    }
    \label{fig:exp_ablation}
\end{figure}

\subsection{Comparative Evaluations in Simulation}
\label{sec:sim_exp}

First, on the \emph{Sorting without Goal Regions} task, we compared our OCP planner against all three re-implemented baselines in Sec.~\ref{sec:baselines}.
We used the same setup as in~\cite{song2020multi}: All movable objects are cubes of size $2.5cm$; the workspace is a square region of size $50cm$; and we evaluated on $6$ different scenes, combinations of $L = 2, 3, 4$ (the number of object classes) and $N = 20, 30$ (the total number of objects).
For all evaluated planners including ours,
we ran experiments with 100 trials on each scene, where the objects were randomly placed at the beginning of each trial.
We set different planning time budgets based on the difficulty of each scene: 
for scenes with $N = 20$, the time limits are $30$, $45$, and $60$ seconds for $L = 2$, $3$, and $4$, respectively; for scenes with $N = 30$, the limits are $90$, $120$, and $180$ seconds. 
A trial is considered a failure if no solution is found within the allotted time of planning.
We report the success rate, planning time, and average number of actions in Fig.~\ref{tab:eval_sim_1}.
As can be seen from the results,
benefiting from our heuristics-guided and more task-relevant motion exploration,
our object-centric planner consistently outperforms all re-implemented baselines across most metrics,
particularly as task complexity increases.
As the number of classes and objects grows, the success rates of all baselines drop significantly.
For instance, without explicit goal-guided action sampling, ILS failed in nearly all trials at $L=4$, $N=4$, 
and H-RRT struggled in every scene with $L > 2$,
while our planner remained consistently successful.

Note that in Fig.~\ref{tab:eval_sim_1}, although we did not re-implement MCTS, we still reported its performance (specifically, success rate and number of actions) using the original results from~\cite{song2020multi} as a reference.
The original MCTS implementation employed a specially designed three-finger robot pusher, which enables the pusher to simultaneously contact multiple objects and concurrently manipulate them by sweeping-like actions.
However, even with a learned policy trained on a large dataset,
MCTS generally requires more actions than our planner to complete the task,
whereas our planner does not need any data for training.
In addition, despite using a parallelized implementation with $8$ threads on an Intel i7-7820X CPU,
MCTS had an average planning time of $2.16s$ per action, while our planner only required an average of $0.5s$ per action on a single thread.

\begin{figure*}[t]
\setlength{\tabcolsep}{8pt}
\centering
\footnotesize
\begin{threeparttable}
 \begin{tabular}{c c | c || c c c c c}
    \hline
    \multicolumn{2}{c|}{Scene} & \multirow{2}{*}{Metric} & \multirow{2}{*}{MCTS~\cite{song2020multi}\tnote{*}} & \multirow{2}{*}{H-RRT~\cite{king2016rearrangement}} & \multirow{2}{*}{ILS~\cite{huang2019large}} & \multirow{2}{*}{kdRRF~\cite{ren2023kinodynamic}} & \multirow{2}{*}{OCP (Ours)} \\
    \# objects & \# classes & & & & & & \\
    \hline
    \multirow{9}{*}{$N = 20$} & \multirow{3}{*}{$L = 2$} & Success Rate & \textbf{100 \%} & 95 \% & 96 \% & 99 \% & \textbf{100 \%} \\
    & & Planning Time (seconds) & -- & 15.5 $\pm$ 5.2 & \textbf{5.9 $\pm$ 2.5} & 16.1 $\pm$ 4.7 & 8.5 $\pm$ 3.1 \\ 
    & & Num. Actions & 36.1 $\pm$ 1.3 & 132.3 $\pm$ 44.2 & 195.1 $\pm$ 55.0 & 30.8 $\pm$ 8.8 & \textbf{28.7 $\pm$ 9.1} \\ 
    \cline{3-8}
    & \multirow{3}{*}{$L = 3$} & Success Rate & 98 \% & 47 \% & 83 \% & 87 \% & \textbf{100 \%} \\
    & & Planning Time (seconds) & -- & 32.7 $\pm$ 7.6 & 16.8 $\pm$ 6.8 & 31.3 $\pm$ 7.1 & \textbf{11.9 $\pm$ 3.7} \\ 
    & & Num. Actions & 66.6 $\pm$ 2.3 & 279.2 $\pm$ 65.0 & 364.9 $\pm$ 98.7 & 56.7 $\pm$ 12.8 & \textbf{41.4 $\pm$ 13.6} \\ 
    \cline{3-8}
    & \multirow{3}{*}{$L = 4$} & Success Rate & 97 \% & 4 \% & 66 \% & 30 \% & \textbf{100 \%} \\
    & & Planning Time (seconds) & -- & 43.7 $\pm$ 3.9 & 38.9 $\pm$ 12.6 & 45.5 $\pm$ 7.3 & \textbf{20.4 $\pm$ 8.3} \\ 
    & & Num. Actions & 80.1 $\pm$ 2.3 & 376.8 $\pm$ 127.8 & 623.6 $\pm$ 151.6 & 77.0 $\pm$ 12.5 & \textbf{63.9 $\pm$ 24.0} \\ 
    \hline
    \multirow{9}{*}{$N = 30$} & \multirow{3}{*}{$L = 2$} & Success Rate & 96 \% & \textbf{98 \%} & 88 \% & \textbf{98 \%} & \textbf{98 \%} \\
    & & Planning Time (seconds) & -- & 47.3 $\pm$ 13.7 & \textbf{19.7 $\pm$ 13.3} & 41.8 $\pm$ 11.7 & 33.9 $\pm$ 14.7 \\
    & & Num. Actions & 77.5 $\pm$ 3.6 & 279.2 $\pm$ 79.9 & 376.7 $\pm$ 96.4 & \textbf{56.8 $\pm$ 15.2} & 69.0 $\pm$ 24.2 \\ 
    \cline{3-8}
    & \multirow{3}{*}{$ L = 3$} & Success Rate & 91 \% & 11 \% & 62 \% & 83 \% & \textbf{97 \%} \\
    & & Planning Time (seconds) & -- & 96.6 $\pm$ 19.4 & 63.2 $\pm$ 26.7 & 81.1 $\pm$ 19.0 & \textbf{52.8 $\pm$ 21.8} \\ 
    & & Num. Actions & 131.5 $\pm$ 5.1 & 573.9 $\pm$ 118.4 & 797.9 $\pm$ 189.6 & 107.9 $\pm$ 24.8 & \textbf{98.8 $\pm$ 31.1} \\ 
    \cline{3-8}
    & \multirow{3}{*}{$L = 4$} & Success Rate & \textbf{89 \%} & 0 \% & 2 \% & 10 \% & 87 \% \\
    & & Planning Time (seconds) & -- & $>$ 180 & \textbf{89.6 $\pm$ 17.6} & 136.8 $\pm$ 34.7 & 91.2 $\pm$ 28.6 \\ 
    & & Num. Actions & 162.6 $\pm$ 4.6 & $>$ 1055 & 1208.5 $\pm$ 64.5 & 174.3 $\pm$ 44.4 & \textbf{155.0 $\pm$ 36.5} \\
    \hline
\end{tabular}
\begin{tablenotes}
    \item[*] The results for MCTS are reported as originally published in~\cite{song2020multi} without re-implementation.
\end{tablenotes}
\end{threeparttable}
\caption{The success rate, planning time, and the average number of actions for different methods evaluated on the \emph{Sorting without Goal Regions} task, with different numbers of objects ($N$) and classes ($L$) in simulation.
}
\label{tab:eval_sim_1}
\end{figure*}

\begin{figure}[t]
\setlength{\tabcolsep}{5pt}
\centering
\footnotesize
\begin{subtable}[h]{\linewidth}
    \setlength{\tabcolsep}{5.5pt}
    \centering
    \caption{Singulate ($N = 33$)}
    \begin{tabular}{c|| c c c c}
        \hline
        Metric & H-RRT~\cite{king2016rearrangement} & ILS~\cite{huang2019large} & kdRRF~\cite{ren2023kinodynamic} & OCP (Ours)\\
        \hline
        S. R. & 10 \% & \textbf{98 \%}  & 71 \% & 97 \% \\
        P. T. & 10.3 $\pm$ 5.8 & 5.9 $\pm$ 4.1 & 8.0 $\pm$ 5.4 & \textbf{2.4 $\pm$ 3.2}\\
        N. A. & 109.1 $\pm$ 61.7 & 109.6 $\pm$ 72.5 & 128.9 $\pm$ 85.5 & \textbf{26.5 $\pm$ 14.4} \\
        \hline
    \end{tabular}
\end{subtable}

\begin{subtable}[h]{\linewidth}
    \setlength{\tabcolsep}{6.4pt}
    \centering
    \caption{Separate ($N = 25$)}
    \begin{tabular}{c|| c c c c}
        \hline
        Metric & H-RRT~\cite{king2016rearrangement} & ILS~\cite{huang2019large} & kdRRF~\cite{ren2023kinodynamic} & OCP (Ours)\\
        \hline
        S. R. & 0 \% & 93 \% & 0 \% & \textbf{98 \%} \\
        P. T. & $>$ 20.0 & 13.8 $\pm$ 2.3 & $>$ 20.0 & \textbf{4.5 $\pm$ 4.2}\\
        N. A. & $>$ 268 & 200.6 $\pm$ 27.5 & $>$ 330 & \textbf{72.4 $\pm$ 47.0} \\
        \hline
    \end{tabular}
\end{subtable}

\begin{subtable}[h]{\linewidth}
    \setlength{\tabcolsep}{4.3pt}
    \centering
    \caption{Sorting with Goal Regions ($L = 4$, $N = 24$)}
    \begin{tabular}{c|| c c c c}
        \hline
        Metric & H-RRT~\cite{king2016rearrangement} & ILS~\cite{huang2019large} & kdRRF~\cite{ren2023kinodynamic} & OCP (Ours) \\
        \hline
        S. R. & 45 \% & 99 \% & 99 \% & \textbf{100 \%}\\
        P. T. & 21.7 $\pm$ 6.0 & 13.4 $\pm$ 4.6 & 13.5 $\pm$ 4.7 & \textbf{4.3 $\pm$ 1.5}\\
        N. A. & 304.2 $\pm$ 82.6 & 402.9 $\pm$ 134.7 & 304.5 $\pm$ 105.7 & \textbf{117.5 $\pm$ 26.1} \\
        \hline
    \end{tabular}
\end{subtable}

\begin{subtable}[h]{\linewidth}
    \setlength{\tabcolsep}{3.5pt}
    \centering
    \caption{Sorting with Goal Regions ($L = 4$, $N = 100$)}
    \begin{tabular}{c || c c c c}
        \hline
        Metric & H-RRT~\cite{king2016rearrangement} & ILS~\cite{huang2019large} & kdRRF~\cite{ren2023kinodynamic} & OCP (Ours) \\
        \hline
        S. R. & 96 \% & \textbf{100 \%} & 93 \% & \textbf{100 \%}\\
        P. T. & 126.6 $\pm$ 23.1 & 142.0 $\pm$ 25.3 & 126.2 $\pm$ 22.4 & \textbf{100.8 $\pm$ 40.5}\\
        N. A. & 464.3 $\pm$ 83.8 & 1017.2 $\pm$ 174.2 & 1091.4 $\pm$ 185.4 & \textbf{374.1 $\pm$ 94.4} \\
        \hline
    \end{tabular}
\end{subtable}

\caption{Simulation evaluations of different methods on \emph{Singulate}, \emph{Separate}, and \emph{Sorting with Goal Regions} ($N = 24$ and $100$) tasks. 
The metrics ``S. R.'', ``P. T.'', and ``N. A." stand for Success Rate, Planning Time (seconds), and Number of Actions, respectively.}
\label{tab:eval_sim_2}
\end{figure}

Next, on the other three tasks, \emph{Singulate}, \emph{Separate}, and \emph{Sorting with Goal Regions} (including two scenes with $N = 24$ and $100$ objects of $4$ classes), 
we compared our OCP planner against all three re-implemented baselines.
For each evaluated task and scene, 
we followed the same setup and time budget as in~\cite{huang2019large} for a fair comparison: 
The workspace was a square region of size $40cm$ (expanded to $125cm$ for the $N=100$ sorting task), and all objects were cubes of size $4cm$;
the time budget was set to $20 s$ for \emph{Singlute} and \emph{Separate} tasks, and $30s$ and $600s$ for \emph{Sorting with Goal Regions} with $N=24$ and $N=100$ respectively.
Also, the sizes of goal regions were also kept the same as in~\cite{huang2019large}.
We conducted $100$ trials for each evaluated planner on each task scene,
and summarized the relevant metrics in Fig.~\ref{tab:eval_sim_2}.
From the results reported in Fig.~\ref{tab:eval_sim_2},
H-RRT and kdRRF baselines performed poorly on the \emph{Separate} task.
This is because, 
unlike our method which generates real-time actions, 
the baselines, although equipped with replanning, did not close the loop tightly enough, 
making them ineffective at precisely pushing objects into small-sized goal regions when discrepancies between modeling and actual execution persist.
More importantly, 
across all tasks,
our OCP planner achieved comparable success rate to ILS but with significantly lower planning time and fewer actions, 
regardless of the task setup or the number of objects.
This highlights a clear improvement in planning efficiency enabled by the object-centric action generation strategy of our planner.

\subsection{Comparative Evaluations on a Physical Robot}
\label{sec:real_world_exp}

\begin{figure}[t]
    \centering
    \hspace{-35pt}\includegraphics[width=1.05\columnwidth]{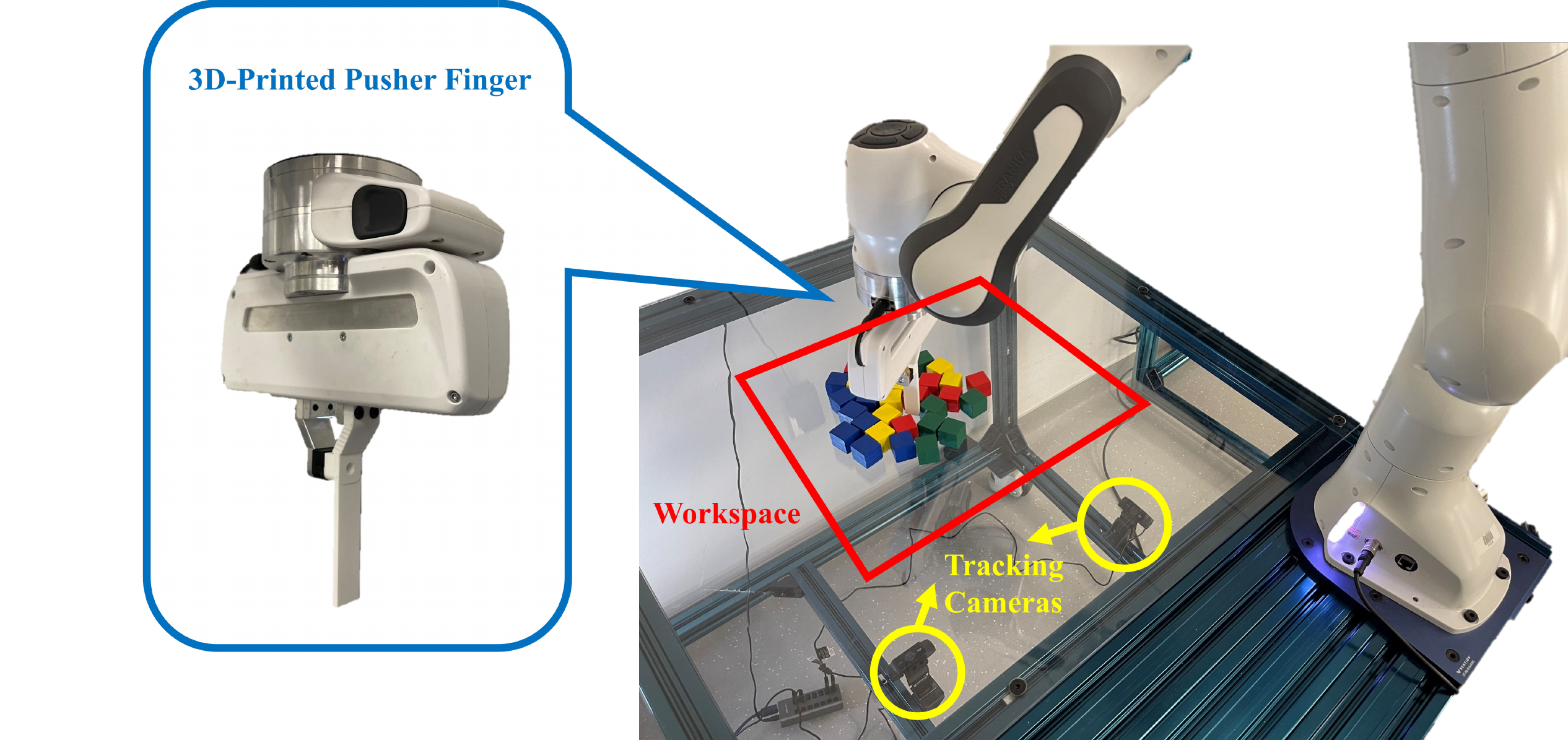}
    \caption{The experiment setup of the physical robot platform.
    The objects were tracked by two cameras below the transparent workspace.}
    \label{fig:real_world_setup}
\end{figure}

To more realistically challenge our OCP planner under real-world uncertainties and explore its deployment potential,
we conducted real-world experiments on a physical 7-DoF Franka Emika Panda robot platform, as displayed in Fig.~\ref{fig:real_world_setup}.
We 3D printed a pusher to replace one finger of the robot gripper.
The workspace plane was made of a transparent panel and the objects were tracked via AprilTags~\cite{olson2011apriltag} by two cameras positioned beneath the workspace.

\begin{figure*}[t]
    \setlength{\tabcolsep}{8pt}
    \centering
    \footnotesize
    \begin{minipage}[b]{\linewidth}
        \centering
        \begin{tikzpicture}
            \node[anchor=south west, align=center, text=black] at (2.5, 3.9) {Scene I ($L=2$, $N = 12$)};
            \node[anchor=south west, align=center, text=black] at (11.35, 3.9) {Scene II ($L=3$, $N = 15$)};
            \node[anchor=south west,inner sep=0] at (0,0){\includegraphics[width=0.95\columnwidth]{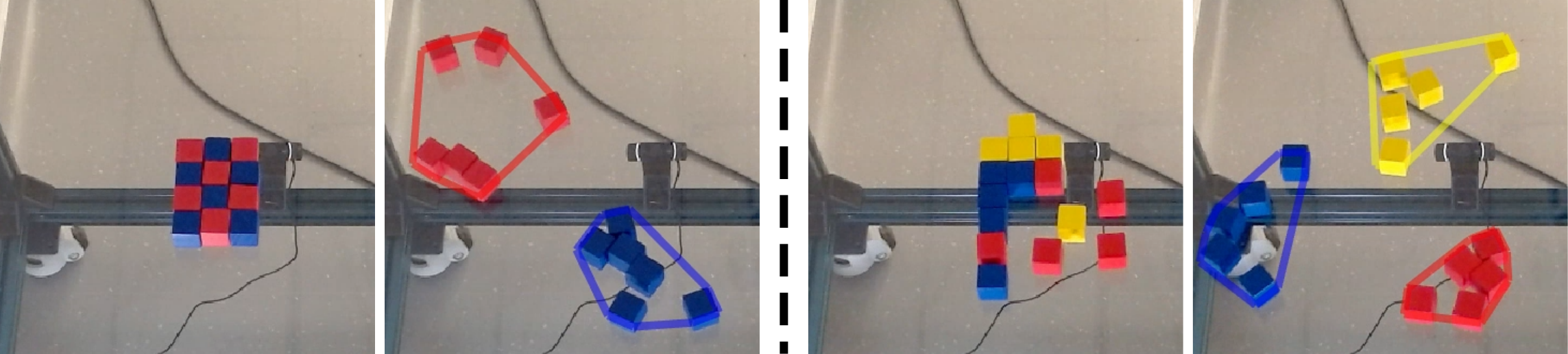}};
            \node[anchor=west, align=left, text=white] at (0.1, 3.7) {Start};
            \node[anchor=west, align=left, text=white] at (4.35, 3.7) {Goal};
            \node[anchor=west, align=left, text=white] at (8.95, 3.7) {Start};
            \node[anchor=west, align=left, text=white] at (13.2, 3.7) {Goal};
        \end{tikzpicture}
        \vspace{5pt}
    \end{minipage}
    
    \begin{minipage}[b]{\linewidth}
        \centering
        \begin{threeparttable}
        \begin{tabular}{c | c || c c c c c}
            \hline
            Scene & Metric & MCTS~\cite{song2020multi}\tnote{*} & H-RRT~\cite{king2016rearrangement} & ILS~\cite{huang2019large} & kdRRF~\cite{ren2023kinodynamic} & OCP (Ours)\\
            \hline
            \multirow{5}{*}{Scene I ($L=2$, $N = 12$)} & Success Rate &  16 / 20 & 6 / 10 & 9 / 10  & \textbf{10 / 10} & \textbf{10 / 10} \\
            & Execution Time (minutes) & -- & 7.6 $\pm$ 1.0 & 6.3 $\pm$ 1.9 & 6.2 $\pm$ 1.9 & \textbf{4.0 $\pm$ 0.5} \\
            & Num. Actions & 37.0 $\pm$ 3.5 & 86.5 $\pm$ 31.3 & 62.3 $\pm$ 20.4 & 23.8 $\pm$ 8.1 & \textbf{19.1 $\pm$ 2.9}\\
            & Planning Time (seconds) & -- & 19.4 $\pm$ 7.8 & \textbf{6.4 $\pm$ 4.0} & 16.5 $\pm$ 5.3 & 9.5 $\pm$ 1.9 \\
            & Plan. Time / Action (seconds) & -- & 0.22 & \textbf{0.15} & 0.69 & 0.50 \\
            \hline
            \multirow{5}{*}{Scene II ($L=3$, $N = 15$)} & Success Rate & 15 / 20  & 0 / 10 & 4 / 10 & 5 / 10 & \textbf{10 / 10}\\
            & Execution Time (minutes) & -- & $>$ 10.0 & 8.1 $\pm$ 2.2 & 6.9 $\pm$ 1.7 & \textbf{3.5 $\pm$ 0.6} \\
            & Num. Actions & 27.7 $\pm$ 2.1 & $>$ 96 & 82.5 $\pm$ 22.4 & 28.2 $\pm$ 8.8 & \textbf{20.5 $\pm$ 4.3}\\
            & Planning Time (seconds) & -- & $>$ 30.1 & 20.4 $\pm$ 19.2 & 27.3 $\pm$ 9.5 & \textbf{9.3 $\pm$ 2.7} \\
            & Plan. Time / Action (seconds) & -- & 0.31 & \textbf{0.21} & 0.97 & 0.46 \\
            \hline
        \end{tabular}
        \begin{tablenotes}
            \item[*] The results for MCTS are reported as originally published in~\cite{song2020multi} without re-implementation.
        \end{tablenotes}
        \end{threeparttable}
    \end{minipage}
    
\caption{Real-world evaluations of different methods on the \emph{Sorting without Goal Regions} task under two different settings. 
}
\label{fig:eval_real_1}
\end{figure*}

First, we evaluated the performance of our planner on tasks without explicit goals.
We used a similar task setup as the physical experiments in~\cite{song2020multi}.
Specifically, the experiments were conducted on the \emph{Sorting without Goal Regions} task, using the same $1$-inch wooden cubes as objects to manipulate.
The task scenes used for evaluation were also the same as~\cite{song2020multi}: 1) $2$ classes $\times$ $6$ objects, and 2) $3$ classes $\times$ $5$ objects.
Since the hardware robot we used is different from~\cite{song2020multi} which had different reachability relative to the workspace, we could not use the exact same workspace as~\cite{song2020multi}.
However, we ensured that our workspace (a rectangle of size $33cm \times 30cm$) had approximately the same area as the one used in~\cite{song2020multi}.
In addition, for a more extensive comparison, 
we re-implemented all the three baselines~\cite{king2016rearrangement, huang2019large, ren2022rearrangement} on the real robot.
Each re-implemented baseline, along with our OCP planner, was evaluated over $10$ trials per task scene,
using the same initial object configuration as in~\cite{song2020multi}, as shown in Fig.~\ref{fig:eval_real_1}.
For both scenes, the time budget for robot execution was set to $10$ minutes, i.e., a task was considered a failure if the robot did not finish the task within $10$ minutes.
The results for relevant metrics are reported in the table of Fig.~\ref{fig:eval_real_1}, where the results for MCTS are directly taken from~\cite{song2020multi} (only the success rate and number of actions are reported).
In the results, our OCP planner always succeeded within $5$ minutes while other baselines could exceed the $10$-minute time budget, particularly in Scene II which involved more objects and classes. 
Additionally, our OCP planner outperformed all baselines regarding the number of actions needed for task completion.
This shows that, in real-world settings, our planner is still consistently able to plan efficiently and generate more effective actions.
In terms of planning time, ILS appeared particularly fast at generating a single action, as it requires only a small number of rollouts in each planning cycle without extensive search or optimization.
However, in real-time scenarios, 
we did not observe substantial differences since all evaluated planners are sufficiently reactive with low per-action planning time.
Execution time and number of actions are usually more representative metrics for assessing practical performance.

\begin{figure}[t]
\setlength{\tabcolsep}{5.5pt}
\centering
\footnotesize
    \begin{minipage}[b]{\linewidth}
    \centering
        \begin{tikzpicture}
            \node[anchor=south west, align=center, text=black] at (2.5, 3.9) {Scene III ($L=4$, $N = 32$)};
            \node[anchor=south west,inner sep=0] at (0,0){\includegraphics[width=0.95\columnwidth]{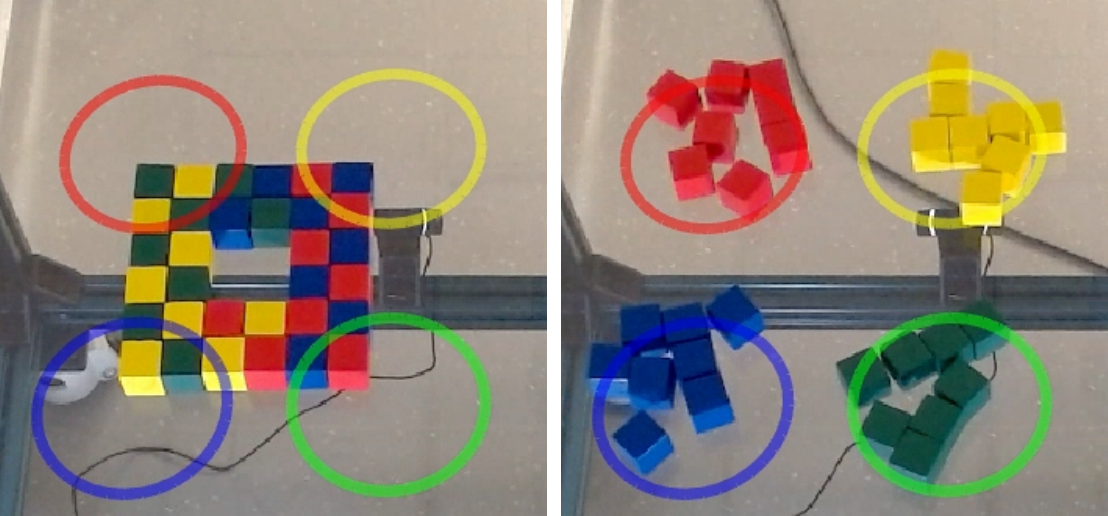}};
            \node[anchor=west, align=left, text=white] at (0.1, 3.7) {Start};
            \node[anchor=west, align=left, text=white] at (4.35, 3.7) {Goal};
         \end{tikzpicture}
        \vspace{10pt}
    \end{minipage}

    \begin{minipage}[b]{\linewidth}
    \centering
    \begin{tabular}{c|| c c c c}
        \hline
        Metric & H-RRT~\cite{king2016rearrangement} & ILS~\cite{huang2019large} & kdRRF~\cite{ren2023kinodynamic} & OCP (Ours)\\
        \hline
        S. R. & 0 / 10 & 0 / 10  & 0 / 10 & \textbf{10 / 10}\\
        E. T. & $>$ 30 & $>$ 30 & $>$ 30 & \textbf{20.7 $\pm$ 4.0} \\
        N. A. & $>$ 242 & $>$ 324 & $>$ 215.8 & \textbf{115.4 $\pm$ 20.1}\\
        P. T. & $>$ 96.8 & $>$ 26.8 & $>$ 181.8 & 83.2 $\pm$ 15.1\\
        P. T. / A. & 0.40 & \textbf{0.08} & 0.84 & 0.72 \\
        \hline
    \end{tabular}
    \vspace{5pt}
    \end{minipage}

    \begin{minipage}[b]{\linewidth}
    \centering
       \input{figs/obj_at_goals.tex}  
    \end{minipage}
\vspace{-15pt}
\caption{Real-world evaluations of different methods on the \emph{Sorting with Goal Regions} task for $4$ classes $\times$ 8 objects, 
where the metrics ``S. R.'', ``E. T.'', ``N. A.'', ``P. T.'', and ``P. T. / A.'' stand for Success Rate, Execution Time (minutes), Number of Actions, Planning Time (seconds), and Planning Time per Action (seconds), respectively.
The bottom figure plots the percentage of objects at their goals throughout the execution of the kdRRF (red), ILS (green), and H-RRT (orange) baselines and our object-centric planner OCP (blue).
The transparent plots show each execution of the $10$ trials, and the solid plots are the average across all $10$ trials.}
\label{tab:eval_real_2}
\end{figure}

Then, we compared our planner against all re-implemented baselines on a task with explicit goal regions,
\emph{Sorting with Goal Regions} as used in~\cite{huang2019large}.
The task involves sorting $N = 32$ cubes (i.e., $4$ classes $\times$ $8$ cubes).
We followed the same experimental setup as~\cite{huang2019large}, 
including cube size, circular goal region size, and a $30$-minute time budget for execution.
Although our workspace dimension ($394mm \times 330mm$) differed from~\cite{huang2019large} due to different experiment space and hardware,
we ensured that the area of our workspace was the same as in~\cite{huang2019large} to challenge our planner with the same packing factor.
We ran $10$ trials for all evaluated planners, 
with the same adversarially designed initial object configuration from~\cite{huang2019large}.
Results are summarized in Fig.~\ref{tab:eval_real_2}.
From the results,
it is clear that our OCP planner consistently enabled the robot with faster task completion than all baselines,
by generating more effective actions in real time.
In Fig.~\ref{tab:eval_real_2} (bottom), for every trial of each evaluated planner, 
we plotted the percentage of objects inside their goal regions throughout the execution.
While all three baselines made steady progress as the execution proceeded, 
it never succeeded within the $30$-minute budget.
In contrast, our planner always successfully sorted all objects across all trials within roughly $25$ minutes of execution.

\begin{figure}[t]
    \centering
    \begin{tikzpicture}
        \node[anchor=south west,inner sep=0] at (0,0){\includegraphics[width=0.95\columnwidth]{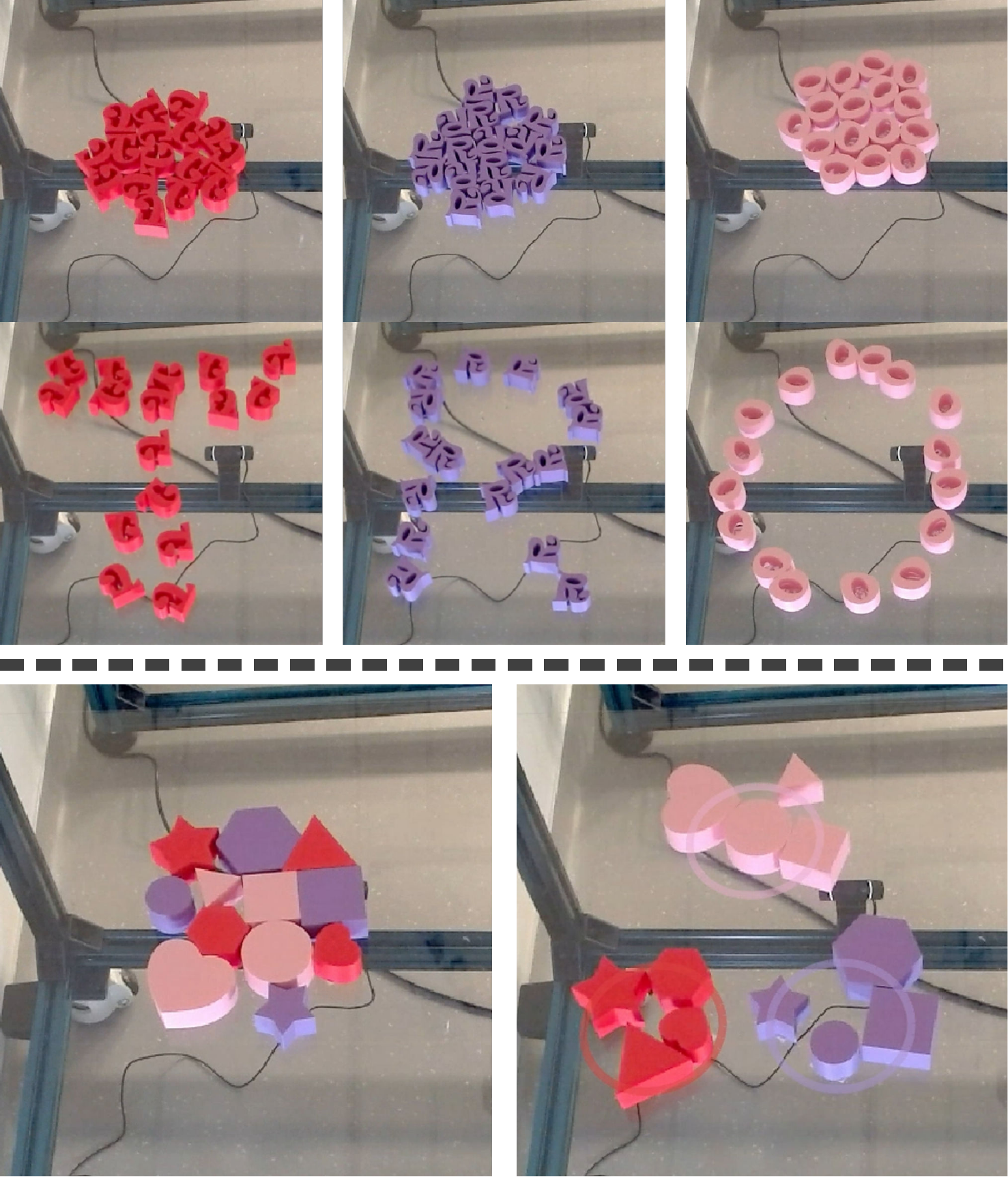}};
    \end{tikzpicture}
    \caption{\emph{Top:} Letter-shaped objects of ``T'', ``R'', and ``O'' are rearranged to form their corresponding letter specified by the goal regions of all objects.
    \emph{Bottom:} Objects of different shapes (heart, star, square, and triangle) and different sizes (ranging from $4cm$ to $7cm$) are sorted into three clusters with different object colors.}
    \label{fig:quality_exp}
\end{figure}

\subsection{Real-world Qualitative Demonstration}

All the aforementioned quantitative evaluations were conducted using cube-shaped objects to ensure fair comparisons with existing baselines.
However, our object-centric planner OCP is not limited to cubes and can generalize to manipulate arbitrary object shapes under real-world settings.
As shown in Fig.~\ref{fig:quality_exp} (top), our planner successfully rearranges non-convex objects shaped like ``T'', ``R'', and ``O'' to draw the corresponding letters formed by their final configurations.
Moreover, in Fig.~\ref{fig:quality_exp} (bottom), we show that our OCP planner can also rearrange objects of different sizes into clusters corresponding to the colors of objects.

\section{Benchmark}
\label{sec:benchmark}

In general, we perceive nonprehensile rearrangement as a combination of global manipulation 
of object clusters and local relocation of individual objects.
To facilitate future research in this area,
we propose a standardized real-world benchmarking protocol to featuring rearrangement tasks that require both intra-class global rearrangement and inter-class relocation.
We also include evaluated performance metrics of our OCP planner as a reference baseline for the benchmark.
All tasks in the benchmark use $1$-inch wooden cubes (Item $\# 65$ in the YCB dataset~\cite{calli2017yale}) as the objects to be rearranged.
The following metrics are used to evaluate relevant methods:
\begin{enumerate}
    \item \emph{Success Rate:} Percentage of successful trials completed within the given execution time budget.
     \item \emph{Execution Time:} Average time of the entire robot execution for successful trials.
    \item \emph{Number of Actions:} Average number of robot actions needed to complete the task successfully.
    \item \emph{Planning Time:} Average total planning time for successful trials, excluding robot execution time.
    \item \emph{Planning Time per Action:} Average runtime the planner takes to generate a valid robot action.
\end{enumerate}

\begin{figure}[t]
\setlength{\tabcolsep}{5pt}
\centering
\footnotesize

\begin{minipage}[b]{\linewidth}
    \centering
    \begin{tikzpicture}
        \node[anchor=south west, align=center, text=black] at (2.5, 3.9) {Scene IV ($L=2$, $N = 10$)};
        \node[anchor=south west,inner sep=0] at (0,0){\includegraphics[width=0.95\columnwidth]{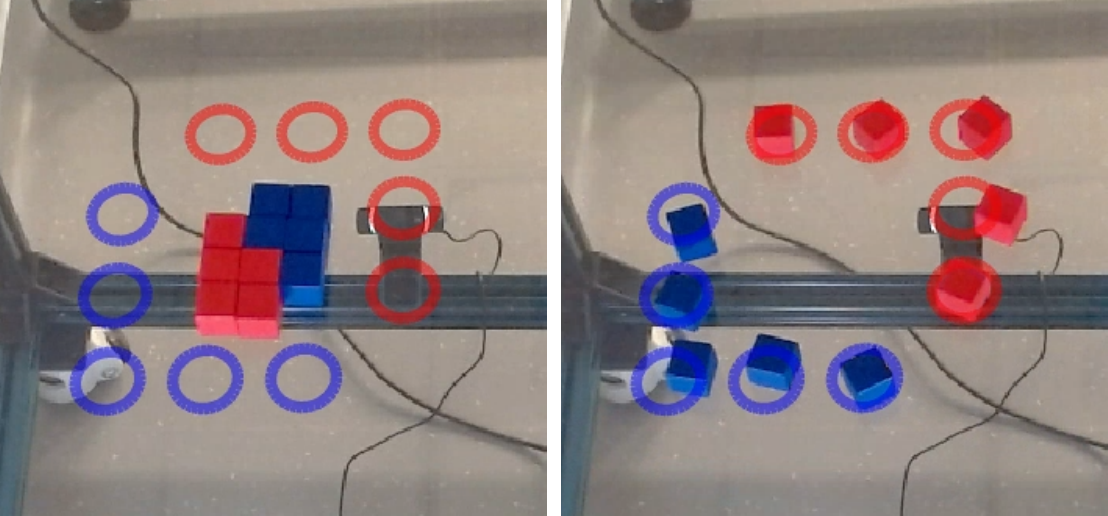}};
        \node[anchor=west, align=left, text=white] at (0.1, 3.7) {Start};
        \node[anchor=west, align=left, text=white] at (4.35, 3.7) {Goal};
    \end{tikzpicture}
    \vspace{5pt}
\end{minipage}

\begin{minipage}[b]{\linewidth}
    \centering
    \begin{tabular}{c|| c | c }
        \hline
        \multirow{2}{*}{Metric} & \multicolumn{2}{c}{Scene}\\
        \cline{2-3}
         & $L = 2$, $N = 10$ & $L = 4$, $N = 16$\\
         \hline
         Time Budget (minutes) & 10 & 15 \\
         \hline
        Success Rate & 10 / 10  & 10 / 10\\
        Execution Time (minutes) & 4.1 $\pm$ 0.9 & 9.6 $\pm$ 1.4 \\
        Num. Actions & 26.9 $\pm$ 4.7 &  51.1 $\pm$ 5.6\\
        Planning Time (seconds) & 6.4 $\pm$ 1.7 & 20.9 $\pm$ 1.8\\
        Plan. Time / Action (seconds) & 0.24 & 0.41 \\
        \hline
    \end{tabular}
\vspace{-5pt}
\end{minipage}

\begin{minipage}[b]{\linewidth}
    \centering
    \vspace{10pt}
    \begin{tikzpicture}
        \node[anchor=south west, align=center, text=black] at (2.5, 3.9) {Scene V ($L=3$, $N = 15$)};
        \node[anchor=south west,inner sep=0] at (0,0){\includegraphics[width=0.95\columnwidth]{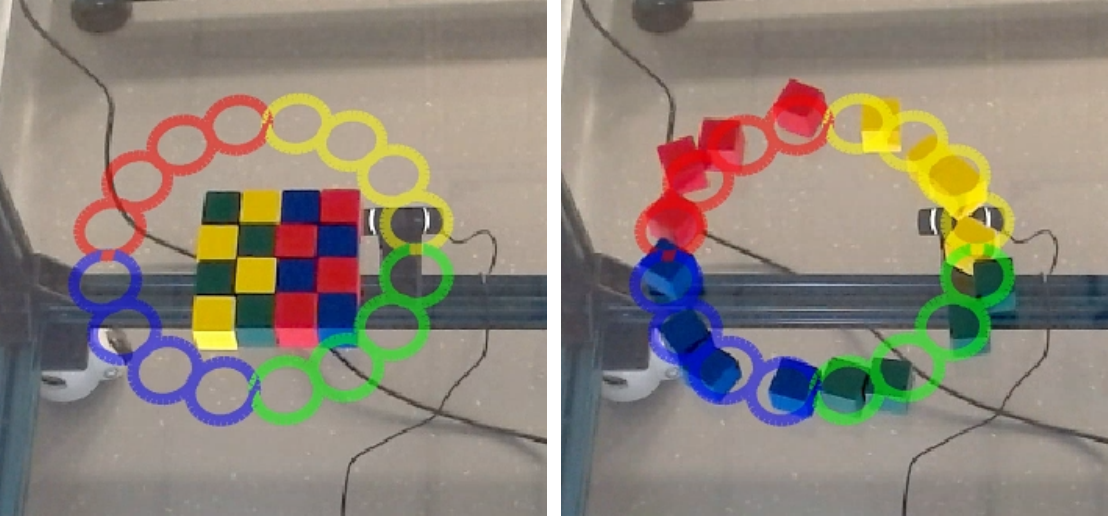}};
        \node[anchor=west, align=left, text=white] at (0.1, 3.7) {Start};
        \node[anchor=west, align=left, text=white] at (4.35, 3.7) {Goal};
    \end{tikzpicture}
\end{minipage}

\caption{Initial and final configurations of the customized Scene IV (top) and Scene V (bottom) in our proposed benchmark.
The relevant metrics evaluated with our proposed object-centric planner OCP on both scenes are given in the table (middle).}
\label{tab:eval_real_3}
\end{figure}

The benchmark includes five scenes, each designed with a different packing factor (i.e., the ratio of the total object area to the workspace area) to reflect varying levels of difficulty.
Each scene also has a virtual out-of-bounds region with a thickness of $2cm$.
The first three scenes, Scene I (Fig.~\ref{fig:eval_real_1} left), II (Fig.~\ref{fig:eval_real_1} right), and III (Fig.~\ref{tab:eval_real_2}), corresponds to those used in the real-world evaluations presented in Sec.~\ref{sec:real_world_exp}.
In addition, 
we introduce another two custom real-world benchmark scenarios,
Scene IV and V, as shown in Fig.~\ref{tab:eval_real_3}:
In both Scene IV and Scene V,
objects of different classes need to be separated into individual clusters, and at the same time, the position of each object in the same class has to lie in one of the distinct non-overlapping goal regions.
Note that the objects in the same class are not unique;
their goal regions are interchangeable and dynamically assigned by minimizing the summed distance between objects and their assigned goals.
Each scene uses a $30cm \times 30cm$ square workspace, and all goal regions are circular with a radius of $2cm$.
We adversarially designed the initial object configurations for both scenes to make them sufficiently challenging.
We ran our OCP planner $10$ times on each scene using execution time budgets of $10$ minutes for Scene IV and $15$ minutes for Scene V.
The statistics of relevant metrics are reported in Fig.~\ref{tab:eval_real_3}.
All five benchmark scenes are summarized below; 
for Scene I, II, and III, the initial and final object configurations, along with associated metrics evaluated using our planner can be found in the referred figures.
\begin{enumerate}
    \item Scene I (packing factor: $0.10$): Sorting $2$ classes $\times$ $6$ objects without explicit goal definitions, as shown in Fig.~\ref{fig:eval_real_1} (left).

    \item Scene II (packing factor: $0.13$): Sorting $3$ classes $\times$ $5$ objects without explicit goal definitions, as shown in Fig.~\ref{fig:eval_real_1} (right).

    \item Scene III (packing factor: $0.20$): Sorting $4$ classes $\times$ $8$ objects with explicit goal definitions, as shown in Fig.~\ref{tab:eval_real_2}.
    
    \item Scene IV (packing factor: $0.095$): Rearranging $2$ classes $\times$ $5$ objects ($N = 10$). 
    The centers of goal regions are at $(9, 9) cm$, $(3, 9) cm$, $(-3, 9) cm$, $(9, 3) cm$, and $(9, -3) cm$ for the first class (blue); and are at $(-9, -9) cm$, $(-9, -3) cm$, $(-9, 3) cm$, $(-3, -9) cm$, $(3, -9) cm$ for the second class (red), as shown in Fig.~\ref{tab:eval_real_3} (top).

    \item Scene V (packing factor: $0.15$): Rearranging $4$ classes $\times$ $4$ objects ($N = 16$).
    The centers of goal regions are evenly spaced along a circle of radius $10cm$, 
    with adjacent goal regions belonging to the same class,
    as shown in Fig.~\ref{tab:eval_real_3} (bottom).
\end{enumerate}

\section{Conclusion}
\label{sec:conclusion}

In this work, 
we proposed a unified framework for nonprehensile rearrangement of multiple movable objects,
by adopting an \emph{object-centric} planning paradigm:
it first reasons about the desired outcomes (i.e., the desired object motions) of robot actions and then realizes the desired object motions via closed-loop pushing actions generated online.
The proposed framework is built on sampling-based kinodynamic planning and is equipped with two complementary exploration modes that alternate during planning.
This design enables both goal-oriented and broad exploration over the problem space,
allowing the robot to more efficiently find desired object arrangements achievable by the robot execution.
Furthermore, 
by interleaving planning and execution, 
the planner can react in real time to perception and modeling uncertainties, 
improving robustness in real-world scenarios.
As a result, our approach can address a range of large-scale, long-horizon, and physics-intensive rearrangement tasks.

With extensive simulation and real-world experiments by comparing against selected state-of-the-art baselines on various rearrangement tasks,
we show that our object-centric planner can improve the planning efficiency by reducing the required runtime, and generate more effective robot actions to reduce the execution time for task completion.
In addition, we propose a real-world benchmarking protocol and provide relevant metrics evaluated using our proposed planner to facilitate future research in nonprehensile rearrangement.

In future work, 
we plan to incorporate prehensile or other nonprehensile primitives other than pushing to generate more diverse and complete rearrangement solutions, while reducing the involved uncertainties during planning.
We also consider generalizing the current framework to more challenging task setups, such as dual-arm manipulation
and rearranging objects with more dynamic motions (e.g., rolling balls).

\bibliographystyle{IEEEtran}
\bibliography{refs}

\end{document}